\documentclass[sigconf,natbib=false]{acmart}
\AtBeginDocument{%
  }

\RequirePackage[
  datamodel=acmdatamodel,
  style=acmnumeric,
  ]{biblatex}

\usepackage{hyperref}
\usepackage{footmisc}
\usepackage{graphicx}
\usepackage{amsmath}
\usepackage{multirow}
\usepackage{tabularx}
\usepackage{cprotect}
\usepackage{verbatimbox}

\newlength{\msize}
\newcommand{\inkimg}[3]{
  \begin{minipage}{#3}{
  \setlength{\msize}{\dimexpr#3-6pt\relax}
  \setlength{\fboxrule}{.1pt}
      \fbox{\includegraphics[width=\msize]{#1}}
      \vspace{-15pt}
      \begin{center}
       \begin{scriptsize}
         #2
       \end{scriptsize}
      \end{center}
  }
  \end{minipage}
}
\newcommand{\train}{\texttt{train}}
\newcommand{\valid}{\texttt{valid}}
\newcommand{\test}{\texttt{test}}
\newcommand{\symbols}{\texttt{symbols}}
\newcommand{\synthetic}{\texttt{synthetic}}
\begin{document}

\title{MathWriting: A Dataset For Handwritten Mathematical Expression Recognition}

\author{Philippe Gervais}
\email{pgervais@acm.org}
\authornote{Work performed while employed at Google}
\authornote{Both authors contributed equally to this research.}
\affiliation{%
  \institution{Inceptive}
  \country{Switzerland}}

\author{Anastasiia Fadeeva}
\email{fadeich@google.com}
\authornotemark[2]
\affiliation{
  \institution{Google DeepMind}
  \city{Zurich}
  \country{Switzerland}
}

\author{Andrii Maksai}
\email{amaksai@google.com}
\affiliation{
  \institution{Google DeepMind}
  \city{Zurich}
  \country{Switzerland}
}

\renewcommand{\shortauthors}{Gervais et al.}

\begin{abstract}
  Recognition of handwritten mathematical expressions allows to transfer scientific notes into their digital form. It facilitates the sharing, searching, and preservation of scientific information. We introduce MathWriting, the largest online handwritten mathematical expression dataset to date. It consists of \textbf{230k human-written samples} and an additional \textbf{400k synthetic ones}. This dataset can also be used in its rendered form for offline HME recognition. One MathWriting sample consists of a formula written on a touch screen and a corresponding \LaTeX{} expression. We also provide a normalized version of \LaTeX{} expression to simplify the recognition task and enhance the result quality. We provide baseline performance of standard models like OCR and CTC Transformer as well as Vision-Language Models like PaLI on the dataset. The dataset together with an example colab is accessible on Github.
\end{abstract}

\ccsdesc[500]{Applied computing~Online handwriting recognition}
\ccsdesc[300]{Applied computing~Optical character recognition}

\keywords{Mathematical Expression Recognition, Online Handwriting Recognition, Digital Ink}

\received{23 February 2025}

\begin{teaserfigure}
\begin{center}
\inkimg{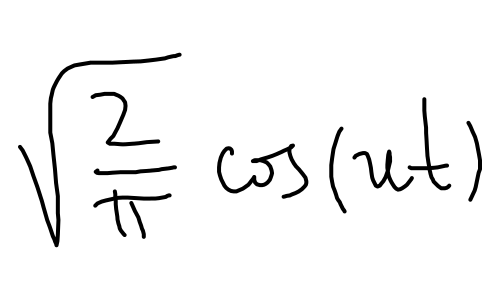}{72a078dfeb8e6027}{100pt}
\inkimg{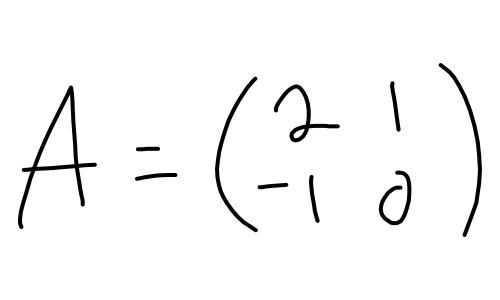}{c96d0c67f82ee512}{100pt}
\inkimg{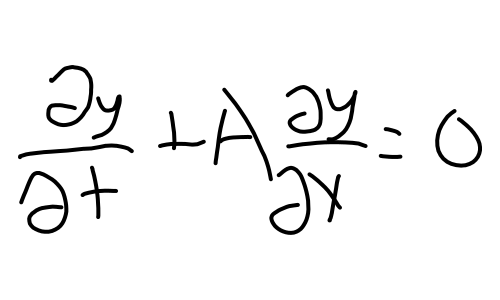}{fe938af2c772a57a}{100pt}
\end{center}
\caption{Three examples of HME from MathWriting. More examples can be found in Appendix~\ref{appendix:ink_examples}. Each ink is accompanied by a unique identifier that matches a corresponding filename in the dataset.
}
\Description{Three handwritten formulas rendered in black on white background.}
\label{fig:ink_examples}
\end{teaserfigure}

\maketitle

\section{Introduction}
\textbf{MathWriting dataset (2.9 GB):} \\
\url{https://storage.googleapis.com/mathwriting_data/mathwriting-2024.tgz} \\
\noindent \textbf{Partial dataset (1.5 MB)}:\\
\url{https://storage.googleapis.com/mathwriting_data/mathwriting-2024-excerpt.tgz} \\

\textbf{Online text recognition} interprets handwritten input on devices with touchscreens like smartphones and tablets. This process analyzes the handwriting as a sequence of strokes, rather than a static image \cite{Graves2009ANC}. The recognition of mathematical expressions (MEs) presents a significant challenge, and has historically received less research attention compared to character and word recognition \cite{ALTaee2020HandwrittenRA}. The fundamental differences between mathematical expression (ME) recognition and text recognition create obstacles to the direct application of improvements developed in one domain to the other. While mathematical expressions (MEs) utilize many of the same symbols as standard text, they are distinguished by a more rigid and two-dimensional structural organization, see Figure~\ref{fig:ink_examples}. Where text can be treated to some extent as a one-dimensional problem amenable to sequence modeling, MEs require two-dimensional analysis due to the importance of symbol spatial positioning. It is also different from symbol segmentation or object detection because the output of a recognizer has to contain the relationship between symbols, serialized in some form (\LaTeX{}, a graph, InkML, etc.). Handwritten mathematical expressions (HMEs), like handwritten text, are more challenging to recognize than their printed counterparts due to increased variability in writing and data scarcity.


Handwritten data acquisition is expensive due to the need for manual input with specialized hardware like touchscreens and digital pens. The publication of the MathWriting dataset aims to address the data requirements for mathematical expression recognition research. Samples include a large number of human-written inks, as well as synthetic ones. MathWriting can readily be used with other online datasets like CROHME~\cite{crohme} or Detexify~\cite{detexifydataset} - we publish the data in InkML format to facilitate this. It can also be used for offline ME recognition by rasterizing the inks, using code provided on the Github page\footnote{\label{Github}\url{https://github.com/google-research/google-research/tree/master/mathwriting}}. 

MathWriting is the largest set of online HME published so far - both human-written and synthetic. It significantly expands the set of symbols covered by CROHME~\cite{crohme}, enabling more sophisticated recognition capabilities. 
Since inks can be rasterized, MathWriting can also been seen as larger than existing offline HME datasets~\cite{deng2017imagetomarkup,yuan2022syntax,aidadataset}.
For these reasons we introduce a new benchmark, applicable to both online and offline ME recognition.

This work's main contributions are:
\begin{itemize}
    \item a large dataset of Handwritten Mathematical Expressions under the Creative Commons Attribution-NonCommercial-ShareAlike 4.0 International\footnote{\url{https://creativecommons.org/licenses/by-nc-sa/4.0/}}.
    \item \LaTeX{} ground truth expressions in normalized form to simplify training and to make evaluation more robust.
    \item Evaluation of different models like CTC Transformer and PaLI on the dataset to show what recognition quality could be achieved with the provided data.
\end{itemize}

The paper focuses on the high-level description of the dataset: creation process, postprocessing, train/test split, ground truth normalization, statistics, and a general discussion of the dataset content to help practitioners understand what can and cannot be achieved with it. All the low-level technical information like file formats can be found in the \verb|readme.md| file present at the root of the dataset archive linked above. We also provide code examples on Github\footref{Github}, to show how to read the various files, process and rasterize the inks, and tokenize the \LaTeX{} ground truth.

\section{Related work}
\label{sec:related_work}
Mathematical expression recognition offers valuable tools for educational software, including features like Math Notes on iPhone \cite{MathNotes}. As tablets such as the Google Pixel, iPad, and Remarkable gain widespread adoption, the demand for effective handwriting tools is growing. 

There are multiple datasets, such as Aida Calculus Math Handwriting Recognition Dataset~\cite{aidadataset} and IM2LATEX-100K~\cite{IM2LATEX}, that focus on image-based (offline) math recognition. 
The Aida dataset consists solely of synthetically generated handwritten math expressions. IM2LATEX-100K has become a key benchmark for assessing performance in typeset math formula recognition. There is also a dataset for typeset math formula detection – TFD-ICDAR 2019 \cite{Mahdavi2019ICDAR2C} that was part of Competition on Recognition of Handwritten Mathematical Expressions (CROHME) 2019. The most widely used dataset for evaluating online math recognition systems originates from the CROHME competition \cite{crohme}. We establish in Section~\ref{sec:comp_crohme} that the MathWriting dataset surpasses CROHME with a richer vocabulary and more unique formulas.

The highest-performing model in the CROHME 2019 competition \cite{Mahdavi2019ICDAR2C} was an LSTM-based encoder-decoder, achieving its result through training on an augmented dataset. This highlights the critical role of training data volume and methods for synthetic data generation. The MyScript model \cite{myscript}, a CTC BLSTM architecture, achieved third place in the competition. Similarly, Section~\ref{sec:experiments} reveals that a CTC Transformer yields strong results on the MathWriting dataset. We also present results obtained with the encoder-decoder model PaLI \cite{chen2023pali}.
\section{Dataset Creation}
MathWriting dataset primarily consists of \LaTeX{} expressions from Wikipedia, more details about the acquisition of expressions are provided in Appendix~\ref{sec:wikipedia_labels}. These expressions were used for both ink collection from human contributors Section~\ref{sec:ink_collection} as well as synthetic data generation Section~\ref{sec:synthetic_samples}. We did a very limited filtering of very noisy human-written examples (described in Appendix~\ref{sec:postprocessing}).   

\subsection{Ink Collection}\label{sec:ink_collection}

Inks were obtained from human contributors through an in-house Android app. Participants agreed to the standard Google terms of use and privacy policy. The task consisted in copying a rendered mathematical expression (prompt) shown on the device's screen using either a digital pen or a finger on a touch screen.
Mathematical expressions used as prompt were first obtained in \LaTeX{} format, then rendered into a bitmap through the \LaTeX{} compiler (see Appendix~\ref{appendix:latex_template} for the template used).
95\% of MathWriting expressions were obtained from Wikipedia. The remaining ones were generated to cover underrepresented cases in Wikipedia, like isolated letters with nested sub/superscripts or complicated fractions (see Section~\ref{sec:wikipedia_labels}). Contributors were hired internally at Google. 6 collection campaigns were run between 2016 and 2019, each lasting between 2 to 3 weeks. Collected data contains only inks and labels, so no personally identifiable information is present in the dataset. Offensive content is highly unlikely because \LaTeX{} expressions were taken from Wikipedia and we conducted a filtering of noisy data (described in Appendix~\ref{sec:postprocessing}).

\subsection{Synthetic Samples and Isolated Symbols}\label{sec:synthetic_samples}

\begin{figure*}[!ht]
    \centering
    \includegraphics[width=300px]{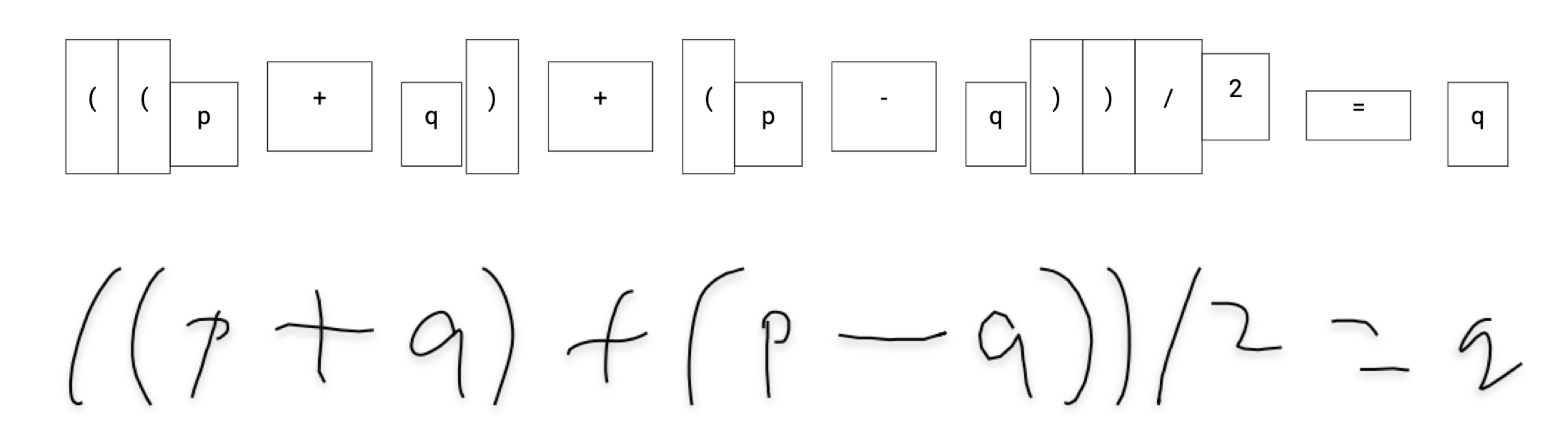}
    \cprotect\caption{An example of a synthetic ink created from bounding boxes with label \verb|((p+q)+(p-q))/2=q|}
    \label{fig:bbox_example}
\end{figure*}

We created synthetic samples in order to further increase the label diversity for training. This also enabled compensating for limitations of the human collection like the maximum length of the expressions, which were limited by the size of the screen they were written on. We used \LaTeX{} expressions from Wikipedia that were not used in the data collection. The resulting \synthetic{} data has a 90th percentile of expression length of 68 characters, compared to 51 in \train{}. This is especially important as deep neural nets often fail to generalize to inputs longer than their training data~\cite{anil2022exploring, Varis_2021}. Using synthetic long inks together with the original human-written inks can help to eliminate that problem as shown in \cite{Timofeev_2023, narayanan2019recognizing}. The synthesis technique is as follows: starting from a raw \LaTeX{} mathematical expression, we computed a DVI file using the \LaTeX{} compiler, from which we extracted bounding boxes. We then used those bounding boxes to place handwritten individual symbols, resulting in a complete expression. See Figure~\ref{fig:bbox_example} for an example of extracted bounding boxes and the resulting synthetic example. We make the bounding box data public to facilitate custom synthetic dataset creation.

Inks for individual symbols are all from the \symbols{} split. They have been manually extracted from inks in \train{}. For each symbol, we manually selected strokes corresponding to it for 20-30 distinct occurrences in \train{}, and used that information to generate a set of individual inks. Similar synthesis techniques have been used by~\cite{crohme} with inks, \cite{deng2017imagetomarkup} and~\cite{aidadataset} with raster images.

The synthetic inks can incorporate symbols from multiple participants which can result in statistical inconsistency. However, even human handwriting displays occasional stylistic variations within one sample (see Appendix~\ref{fig:app_samples} ink $02229a0c174d8dbe$). We argue that having more diversity in synthetic dataset is valuable for training robust recognition models. We also show in ablation experiments (Section~\ref{sec:ablation}) that \synthetic{} data improves the quality of recognition. 

Another important distinction between the synthetic and human-written inks is the stroke order. For synthetic inks, stroke order follows the order of the bounding boxes in the DVI file, which can be different from the usual order of writing for mathematical expressions. However, the writing order within a given symbol is consistent with human writing.

\subsection{Dataset Split}\label{sec:split}

\begin{figure*}[!ht] 
    \centering
    \inkimg{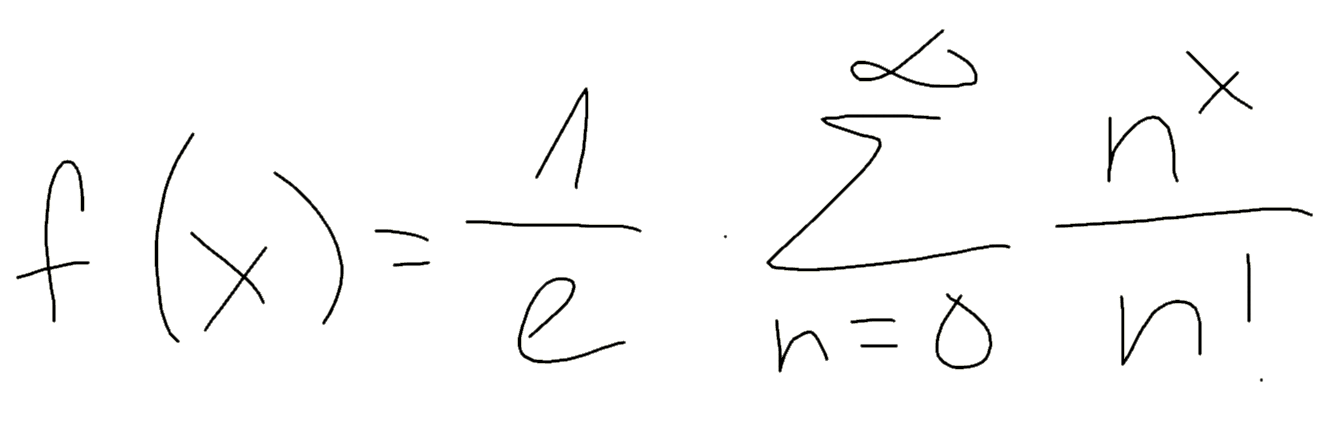}{51b364eb9ba2185a}{250px}

    \cprotect\caption{An example from the \train{} split, with its labels:
    Raw: \verb|f(x)=\frac1e\cdot \sum_{n=0}^{\infty}{n^{x}\over n!}|
    Normalized: \verb|f(x)=\frac{1}{e}\cdot\sum_{n=0}^{\infty}\frac{n^{x}}{n!}|}
    \label{fig:ink-example}
\end{figure*}

MathWriting is composed of five different sets of samples, which we call 'splits': \train{}, \valid{}, \test{}, \symbols{}, and \synthetic{}. The splits \train{}, \valid{} and \test{} consist only of human-written examples. The split \symbols{} is provided for synthetic data generation and is not used in training. The split of human-written samples between \train{}, \valid{} and \test{} was partially done based on writers, partially based on labels. More details are provided in Appendix~\ref{sec:dataset_split}. Experiments have shown that a more important factor than the handwriting style was whether the \textit{label} had already been seen during training. This fact is also supported by research in the area of compositional generalization \cite{keysers2020measuring}. In the published version, \valid{} has a 55\% (8.5k samples) intersection with \train{} based on unique normalized labels, and \test{} has an 8\% intersection (647 samples). We chose to have a low intersection between \train{} and \test{} in order to correctly measure generalization of trained models to unseen labels.

\subsection{Label Normalization}\label{sec:normalization}

All samples in the dataset come with two labels: the \LaTeX{} expression that was used during the data collection (annotation \texttt{label} in the InkML files), and a normalized version of it meant for model training, which is free from a few sources of confusions for an ML model (annotation \texttt{normalizedLabel}). An example with original and normalized labels is provided in Figure~\ref{fig:ink-example}. Label normalization covers three main categories (details are provided in Appendix~\ref{sec:appendix_normalization}):

\begin{itemize}
    \item variations used in print that can't be reproduced in handwriting - e.g. bold, italic - or that haven't been reproduced consistently by contributors.
    \item non-uniqueness of the \LaTeX{} syntax. e.g. \verb"\frac{1}{2}" and \verb"1\over 2" are equivalent.
    \item visual variations that can reproduced in handwriting but can't reliably be inferred by a model. This includes size modifiers like \verb|\left|, \verb|\right|.
\end{itemize}

We provide the raw labels to make it possible to experiment with alternative normalization schemes, which could lead to better outcomes for different applications.

\subsubsection{Limitations of normalization}\label{sec:normalization_limit}

The normalization process is purely syntactic, and can not cover cases where the meaning of the expression has to be taken into account. For example, a lot of expressions from Wikipedia use \verb|cos| instead of \verb|\cos|. It is often clear to a human reader whether the sequence of characters c,o,s represents the \verb"\cos" command or simply three letters. However, this can not be reliably inferred by a syntactic parser, for example in \verb|tacos| vs \verb|ta\cos|. An alternative would be to update the raw labels, which we didn't do because we wanted to keep the information that was used during the collection as untouched as possible. Similarly, cases like \verb|10^{-1}| usually mean \verb|{10}^{-1}|, though they render exactly the same. We made the choice to normalize to the former because it's the only option with a purely syntactic normalizer. It's also better than not removing these extra braces because it gives more consistent label structures, which simplifies the model training problem. 
\section{Dataset Statistics}\label{sec:data_stat}

In this section we describe the key characteristics of MathWriting and compare it to CROHME23 \cite{crohme}. In Table~\ref{table:main_statistics} we provide the information about the volume of the dataset splits both in terms of examples (inks) and unique labels.

\begin{table}[!ht]
\caption{Statistics on different subsets of MathWriting dataset.}
\label{table:main_statistics}
\smallskip
\footnotesize
\centering
\begin{tabular}{lcccc }
\toprule
  & \train & \synthetic & \valid & \test \\
  \midrule
 \# distinct inks & 230k &  396k & 16k & 8k \\
 \# distinct labels & 53k & 396k & 8k & 4k \\
\bottomrule
\end{tabular}
\end{table}

\subsection{Label Statistics}
\begin{figure*}[!ht]
    \centering
    \includegraphics[width=\textwidth]{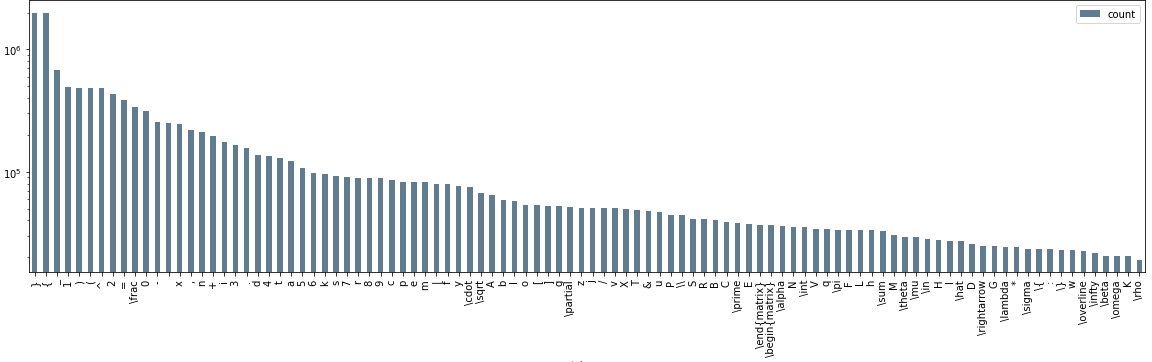}
    \cprotect\caption{Histogram of the top-100 most frequent tokens in MathWriting.}
    \label{fig:top_100_tokens}
\end{figure*}

MathWriting contains 457k unique labels after normalization (see Section~\ref{sec:normalization}). From Table~\ref{table:main_statistics} we see that most unique expressions are covered by the synthetic portion of the dataset. However, the absolute number of unique expressions in human-written part is still high – 61k. This underlines the importance of synthetic data as it allows models to see a much bigger variety of expressions. It is important to note that the \synthetic{} split has essentially no repeated expressions. On the other hand, in real data multiple different writings of the same expression are quite common (see Figure~\ref{fig:frequency_plot} in Appendix~\ref{sec:dataset_additional}). This fact allows us to separately evaluate model's quality on expressions that were observed during training and that those that hadn't. As seen in Table~\ref{table:label_intersection} the biggest intersection in expressions is between \valid{} and \train{}. The minimal overlap between \test{} and \train{} splits is beneficial for assessing a model's ability to generalize to expressions that were not seen in train.



\begin{table}[!ht]
\caption{Counts of unique labels shared between MathWriting splits}
\label{table:label_intersection}
\smallskip
\footnotesize
\centering
\begin{tabular}{lcccc }
\toprule
  & \train & \synthetic & \valid & \test \\
  \midrule
 \train & - & 0 & 3.6k & 355 \\
 \synthetic & 0 & - & 0 & 0 \\
 \valid & 3.6k & 0 & - & 239\\
 \test & 355 & 0 & 239 & -\\
\bottomrule
\end{tabular}
\end{table}

The median length of expressions in characters is~26 which is comparable to one of the most popular English recognition datasets IAMonDB \cite{liwicki2005iam} which has median of 29 characters. However, it is important to note that \LaTeX{} expressions have tokens that span multiple characters like \verb|\frac|. The median length of expressions in tokens (provided in Appendix~\ref{sec:tokens}) is 17, thus making training a model on tokens rather then characters easier due to shorter target lengths \cite{seq_len, neishi-yoshinaga-2019-relation}. We want to emphasize that MathWriting can be used with a different tokenization scheme and token vocabulary from what we propose in Appendix~\ref{sec:tokens}.  In Figure~\ref{fig:top_100_tokens} we show the number of occurrences for the most frequent tokens. Tokens \verb|{| and \verb|}| are by far the most frequent as they are integral to the \LaTeX{} syntax.

\subsection{Ink Statistics}

\begin{table}
\caption{Ink statistics for MathWriting.}
\begin{center}
\begin{tabular}{lccc}
\toprule
                   & 10th percentile    & median & 90th percentile   \\ \midrule
\# strokes & 5 & 14 & 39  \\ 
\# points & 131 & 350 & 1069  \\
writing time (sec) & 1.88 & 6.03 & 16.42  \\
aspect ratio & 1.32 & 3.53 & 9.85 \\
\bottomrule
\end{tabular}
\end{center}
\label{tab:ink_stats}
\end{table}

\begin{figure}[!h]
\begin{center}
\inkimg{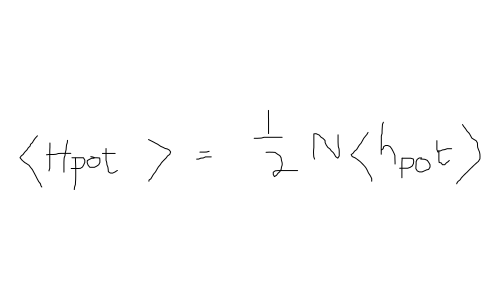}{a21ff5968b022586}{140pt}
\inkimg{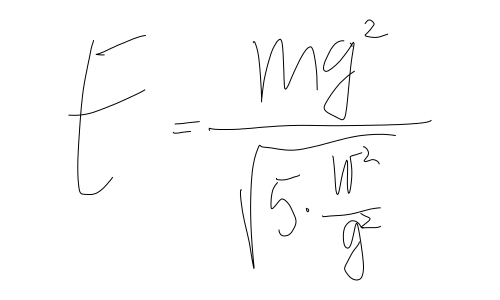}{17b337d8b7ac540b}{140pt}
\end{center}
\cprotect\caption{Top: an ink with very low sampling rate (9.4 points per second) \\
Bottom: an ink with very high sampling rate (260 points per second)
}
\label{fig:natural_sampling}
\end{figure}

Each ink in MathWriting dataset is a sequence of strokes $\text{I} = [s_0, \ldots, s_n]$, each stroke $s_{i}$ consisting of points. A point is represented as a triplet $(x, y, t)$ where $x$ and $y$ are coordinates on the screen and $t$ is a timestamp. In Table~\ref{tab:ink_stats} we provide statistics on number of strokes, points, and duration of writing. It's important to note that as inks were collected on different devices, the absolute coordinate values can vary a lot. In human-written data the time information~$t$ always starts from 0 but it is not always the case in the \synthetic{} split. Different samples often have different sampling rates (number of points written in one second) due to the use of different devices (see Figure~\ref{fig:natural_sampling}). More details in Section~\ref{sec:devices}. Consequently, the same ink written on two different devices can result in inks with a different number of points. For human-written inks, the sampling rate is consistent between strokes, but it is not the case for synthetic ones. In order to accommodate a model and make sequences shorter, inks can be resampled in time.


\subsection{Devices Used}\label{sec:devices}

Around 150 distinct device types have been used by contributors. In most cases inks were written on smartphones using a finger on a touchscreen. However, there are cases where tablets with styluses were used. The main device used in this case is Google Pixelbook, which accounted for 51k inks total (see Table~\ref{tab:devices}, Appendix~\ref{sec:dataset_additional}). Out of all device types, 37 contributed more than 1000 inks. Note that writing on a touchscreen with a finger or a stylus results in different low-level artifacts. All devices were running the same Android application for ink collection, regardless of whether their operating system was Android or ChromeOS.

\subsection{Comparison With CROHME23}
\label{sec:comp_crohme}

In this section we compare main dataset statistics of MathWriting and CROHME23 \cite{crohme} as it is a popular publicly available dataset for HME recognition. In terms of overall size, MathWriting has nearly 3.9~times as many samples and 4.5~times as many distinct labels after normalization, see Table~\ref{tab:mw_cr23_counts}. A significant number of labels can be found in both datasets (47k), but the majority is dataset-specific. This suggests that combining both datasets during training could yield improved HME recognition quality. MathWriting has more human-written inks than CROHME23 as seen in Table~\ref{tab:mw_cr23_human_synthetic}, and contains a much larger variety of tokens. It has 254 distinct tokens including all Latin capital letters and almost the entire Greek alphabet. It also contains matrices, which are not included in CROHME23. Therefore, more scientific fields like quantum mechanics, differential calculus, and linear algebra can be represented using MathWriting. 



\begin{table}
\centering
\begin{minipage}[t]{.5\textwidth}
\cprotect\caption{Counts of inks, distinct labels and distinct tokens used in MathWriting and CROHME23. The single token present in CROHME23 but not in MathWriting is the literal dollar sign \verb|\$|.}
\label{tab:mw_cr23_counts}
\begin{center}
\begin{tabular}{lp{15mm}p{15mm}p{15mm}}
\toprule
       & MathWriting     & CROHME23   & Common \\
       \midrule
Inks   & 650k & 164k & 0   \\ 
Labels & 457k & 102k & 47k \\ 
Vocab & 254    & 105    & 104 \\
\bottomrule
\end{tabular}
\end{center}
\end{minipage}
\hfill
\begin{minipage}[t]{.5\textwidth}
\caption{Count of human-written and synthetic inks for MathWriting and CROHME23. Human-written inks represent 38\% of the total for MathWriting, and 10\% for CROHME23.}
\begin{center}
\begin{tabular}{lp{15mm}p{15mm}}
\toprule
                   & MathWriting     & CROHME23   \\
\midrule
human & 253k & 17k  \\ 
\synthetic     & 396k & 147k \\
\bottomrule
\end{tabular}
\end{center}
\label{tab:mw_cr23_human_synthetic}

\end{minipage}
\end{table}

\section{Experiments} 

\subsection{Evaluation setup}

\begin{table*}[!ht]
\caption{Recognition results for different models. The evaluation metric is reported on both the \valid{} and \test{} splits.}
\label{table:recognition}
\smallskip
\centering
\begin{tabular}{lp{6mm}p{6mm}cccccc } 
\toprule
 &  &  & \multicolumn{3}{c}{\valid} & \multicolumn{3}{c}{\test}\\
 Model & Time & Params & CER $\downarrow$ & EM $\uparrow$ & $\le 1$ dist  $\uparrow$& CER $\downarrow$ & EM  $\uparrow$& $\le 1$ dist $\uparrow$\\
  \midrule
 OCR \cite{GoogleOcr} & no & - & 6.50 & 64 & 76 & 7.17 & 53 & 68\\
 CTC Transformer \cite{fadeeva2024representing} & yes & 35M & 4.52 (0.08) & 71 (0.46) & 81 (0.3) & 5.49 (0.05) & 60 (0.42) & 72 (0.42) \\
 PaLI \cite{chen2023pali} & yes & 700M & 4.47 (0.08) & 76 (0.25) & 83 (0.11) & 5.95 (0.06) & 64 (0.35) & 73 (0.22) \\
 PaLIGemma \cite{beyer2024paligemmaversatile3bvlm} & yes & 3B & 3.95 (0.04) & 80 (0.04) & 86 (0.03) & 5.97 (0.08) & 69 (0.46) & 77 (0.26) \\
 \bottomrule
\end{tabular}
\end{table*}

\begin{figure*}[!ht]
    \centering
    \includegraphics[width=300px]{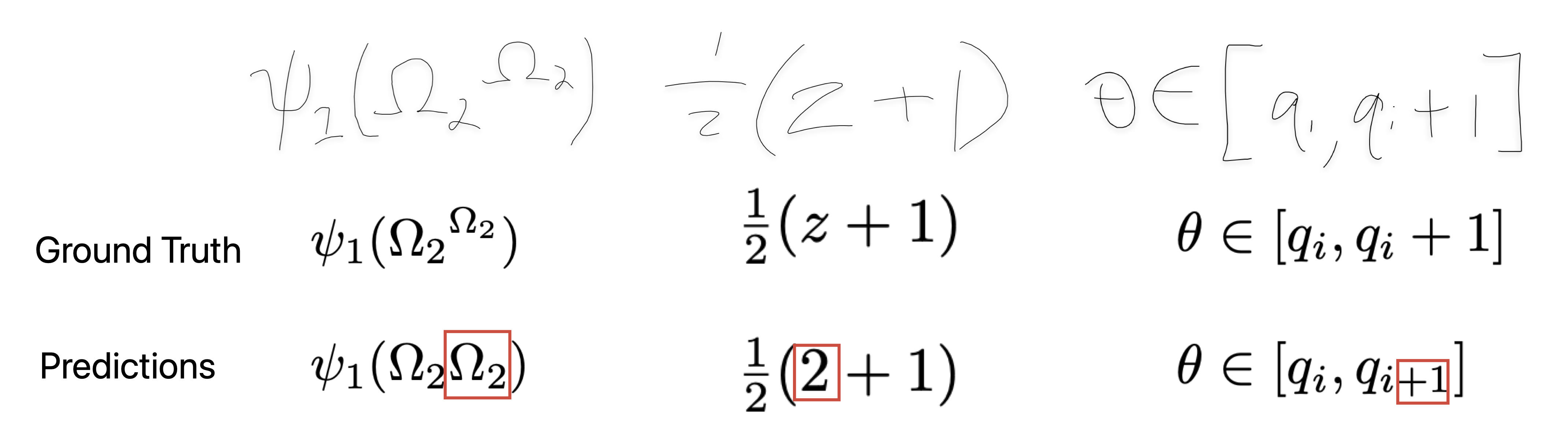}
    \cprotect\caption{Examples of recognition mistakes from the CTC Transformer model. We observe similar mistakes from the other models.}
    \label{fig:recognition_mistakes}
\end{figure*}

We propose the following evaluation setup based on MathWriting for the quality of handwriting math expression recognition.

\begin{itemize}
\item \textbf{evaluation samples}: the \test{} split of MathWriting.

\item \textbf{metric}: character error rate (CER)~\cite{michael2019evaluating}, where a "character" is a \LaTeX{} token as defined by the code in Appendix~\ref{sec:tokenization}.
\end{itemize}

We provide a reference implementation of the evaluation metric at the Github page \footref{Github}. We propose the use of CER as a metric to make results comparable to other recognition tasks like text recognition~\cite{retsinas2024best, kass2022attentionhtr}, and the use of \LaTeX{} tokens instead of ASCII characters so that an error on a single non-latin letter (e.g. \verb|\alpha| recognized as \verb|a|) counts as one instead of many.

\subsection{Baseline Recognition Models} \label{sec:experiments}

Table~\ref{table:recognition} compares different recognition methods discussed in Section~\ref{sec:related_work}. All models are trained exclusively on the MathWriting dataset (\train{} and \synthetic{}), except for the OCR API that was trained on other datasets as well. The following models represent different approaches to handwriting recognition – offline \cite{kass2022attentionhtr}, online \cite{carbune2020fast} and mixed \cite{fadeeva2024representing}. Through fine-tuning, we establish a benchmark for the quality attainable with MathWriting data.

\paragraph{OCR}
This is a publicly available Document AI OCR API~\cite{GoogleOcr}, which processes bitmap images. It has been trained partly on samples from MathWriting. We sent inks rendered with black ink on a white background and searched for optimal image size and stroke width to get the best evaluation result from the model. We also  fine-tuned TrOCR model following methodology outlined in \cite{Rogge_Transformers_Tutorials_2020} and observed only 25\% Exact Match on \valid{} split. Similar behavior was observed with TrOCR \cite{li2022trocrtransformerbasedopticalcharacter} on CROHME dataset \cite{TrOCR_latex} and the possible explanation is that the text tokenizer used by TrOCR is not well adapted to handle LaTeX expressions. 

\paragraph{CTC Transformer}\label{sec:ctc_transformer}
This model is a transformer base with a Connectionist Temporal Classification loss on top (CTC)~\cite{graves2006connect}. It contains 11 transformer layers with an embedding size of 512. We used swish activation function and dropout of 0.15 as those parameters performed best on \valid{}. We train with an Adam optimizer, learning rate of 1e-3, batch size 256 for 100k steps. One training run took 4 hours on 4 TPU v2. We trained from scratch and exclusively on MathWriting (\train{} and \synthetic{}). The model is similar to~\cite{carbune2020fast}, replacing LSTM layers by Transformer layers and not using any external language model on top.

\paragraph{VLM}\label{sec:vlm}
We fine-tuned a large Vision-Language Model PaLI~\cite{chen2023pali} with encoder-decoder architecture on MathWriting (\train{} and \synthetic{}). We used the representation proposed in \cite{fadeeva2024representing} where an ink is represented as both a sequence of points (similar to CTC Transformer) and its rasterized version (similar to OCR). We train three models with different data shuffling for 200k steps with batch size 128, learning rate 0.3 and dropout 0.2. One training run took 14 hours on 16 TPU v5p. Models were fine-tuned exclusively on \train{} and \synthetic{} MathWriting data. Overall, it took 2 TPU v2 days and 28 TPU v5p days to run the experiments.

We also fine-tuned a publicly available PaLIGemma model \cite{beyer2024paligemmaversatile3bvlm} that utilizes a decoder-only LLM – Gemma \cite{gemmateam2024gemmaopenmodelsbased}. This model has recently shown strong performance on captioning and VQA tasks after fine-tuning. The model is trained on image input with 448px resolution. We leverage the image representation with speed information from \cite{fadeeva2024representing}, that demonstrated superior performance compared to black-on-white rendering. We've trained the model with learning rate 1e-4 and batch size 512 on 64 TPU v5p over 36 hours.

Table~\ref{table:recognition} shows the evaluation comparison between the four models. The OCR model has no information about the order of writing and speed (offline recognition), which explains its lower performance than methods that take time information into account (online recognition). The other methods – PaLI, PaLIGemma and CTC Transformer perform significantly better than OCR. These results show that our dataset can be used to train classical recognition models like CTC transformer as well as more recent architectures like VLMs.

Figure~\ref{fig:recognition_mistakes} shows examples of model mistakes. Two of the main causes of mistakes are confusing similar-looking characters like ``z'' and ``2'', and errors in the structural arrangement of the characters, for instance not placing a sub-expression in a subscript or superscript.

\subsection{Synthetic data ablation}
\label{sec:ablation}
In Section~\ref{sec:synthetic_samples} we described the synthetic dataset generation process. This section examines the empirical impact of synthetic data on the model's performance metrics. We have opted for a CTC Transformer model for this purpose as it allows for faster experimentation. Table~\ref{table:ablation_synthetic} shows that there is a 10\% decrease in CER on the \test{} dataset when synthetic data is used in training, and close to 5\% on the \valid{} dataset. These results demonstrate the substantial benefit of synthetic data that we provide for recognition training.

\begin{table}[!ht]
\caption{Ablation of synthetic data on CTC transformer model.}
\label{table:ablation_synthetic}
\smallskip
\footnotesize
\centering
\begin{tabular}{lcc }
\toprule
  \synthetic & CER eval $\downarrow$ & CER test $\downarrow$ \\
  \midrule
 with & \textbf{4.52 (0.08)} & \textbf{5.49 (0.05)} \\
 without & 4.64 (0.07) & 6.2 (0.08) \\
\bottomrule
\end{tabular}
\end{table}

\section{Discussion}

\subsection{Differences in Writing Style}

\begin{figure}
    \begin{center}
      \inkimg{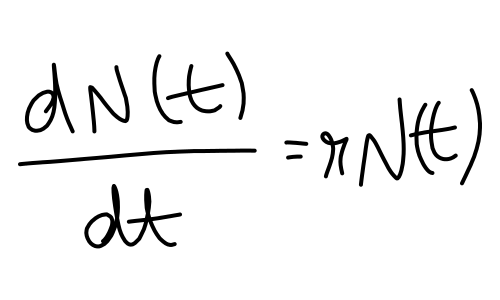}{d4e9c7b8f1ffb958}{120pt}
      \inkimg{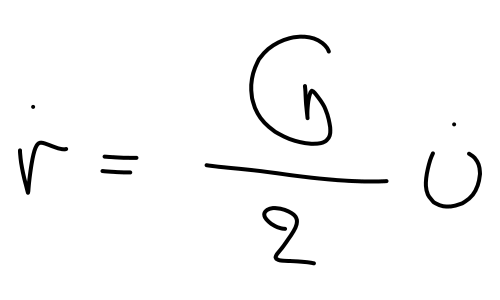}{0dca307a1895512d}{120pt}
      \inkimg{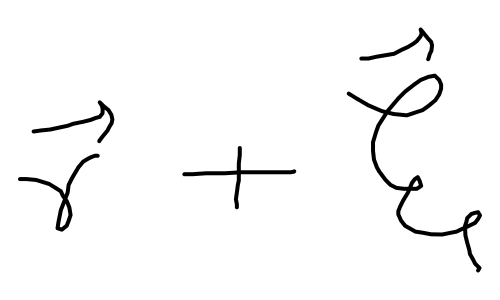}{45af9641d8d2ed56}{120pt}
    \end{center}
    \caption{Three ways of writing a lowercase 'r'.}
    \label{fig:writingstyle_r}
\end{figure}

\begin{figure}
    \centering
    \inkimg{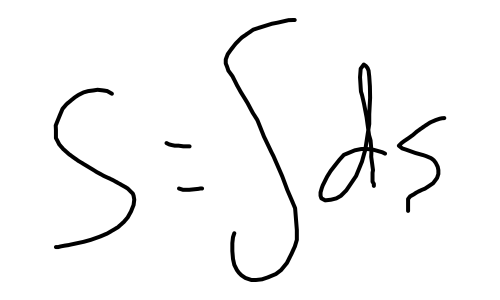}{a933cd67f7891dc8}{130pt}
    \inkimg{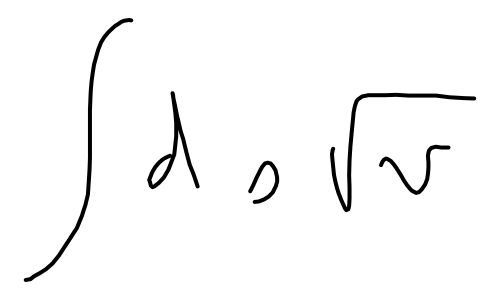}{ecc157b89c3e344d}{130pt}
    \caption{Two ways of writing lowercase s.}
    \label{fig:writingstyle_s}
\end{figure}

The number of contributors was large enough that a variety of writing styles are represented in the dataset. An example for different ways of writing letter `r' can be seen in Figure~\ref{fig:writingstyle_r}. Additional examples are provided in Figure~\ref{fig:writingstyle_s}. Similar though less obvious differences exist for other letters. Style differences also show through writing order (example – Figure~\ref{fig:writingstyle_fractions}, Appendix~\ref{sec:discussion_additional}).

\subsection{Recognition Challenges}\label{sec:rec_challenge}

\begin{figure}
\begin{center}

\inkimg{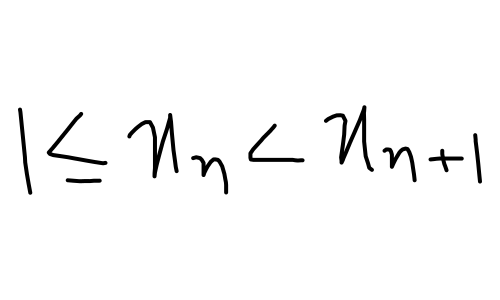}{357313f77d65b804}{130pt}
\inkimg{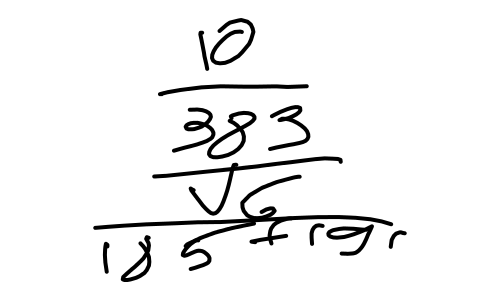}{b8b8cff97f5c044a}{130pt}
\end{center}
\caption{Top: character ambiguity. Is it $1\le x_n < x_{n+1}$ or $1\le n_\eta < n_{\eta+1}$ ?
Bottom: what is the fraction nesting order?}
\label{fig:ambiguity1}
\end{figure}

\begin{figure}
\begin{center}
\inkimg{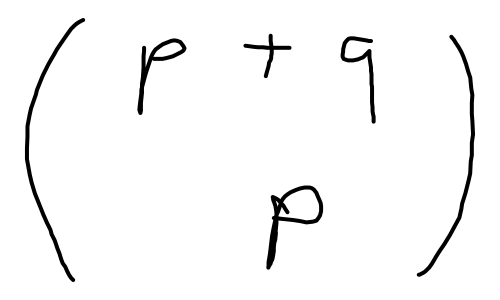}{96bf9fb2da96db9e}{130pt}
\inkimg{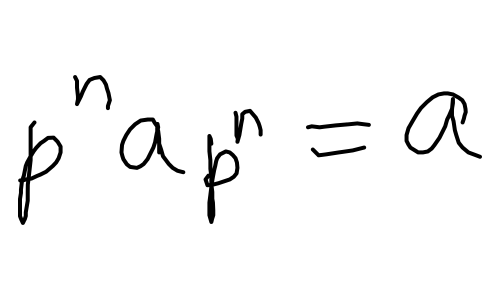}{e50b0275ef2c9549}{130pt}
\end{center}
\cprotect\caption{
Top: \verb|\binom| or 2-element matrix? 
Bottom: $p^n a_{p^n} =a$ or $p^n a_{p}n =a$ ?
}
\label{fig:ambiguity2}
\end{figure}

MathWriting presents some inherent recognition challenges, which are typical of handwritten representations. For example, it's not really possible to distinguish these pairs from the ink alone:\\ \verb|\frac{\underline{a}}{b}| vs \verb|\frac{a}{\overline{b}}|,\\ and \verb|\overline\omega| vs \verb|\varpi|. We'd like to point out that these ambiguities are not an issue for humans in practice, because they rely on contextual information to disambiguate: a particular writing idiosyncrasy, consistency with nearby expressions, knowledge of the scientific domain, etc. See Figures~\ref{fig:ambiguity1} and~\ref{fig:ambiguity2} for more examples.

\subsection{Dataset Applications and Future Work}\label{ref:applications_future}

Mathwriting can be used to train recognizers for a large variety of scientific fields, and is also large enough to enable synthesis of mathematical expressions. Combining it with other large datasets like CROHME23 would increase the variety of samples even further, both in terms of writing style and number of expressions, likely improving the performance of a model.

Bounding box information for synthetic samples together with individual symbols are provided to enable experimentation with synthetic ink generation. Synthetic samples were generated through the straightforward process of pasting inks of individual symbols (\symbols{}) exactly where bounding boxes were located. This gives synthetic samples a very regular structure, see Figure~\ref{fig:bbox_example}. It is possible to improve this process by modifying the location, size or orientation of bounding boxes prior to generating the synthetic inks. This would soften \LaTeX{}'s rigid structure and make synthetic data closer to human handwriting. Another application of these bounding boxes would be to bootstrap a recognizer that would also return character segmentation information. This kind of output is critical for some UI features - for example, editing an handwritten expression.

MathWriting can also be improved by varying the label normalization. Changing it can have different benefits depending on the application, as mentioned above. We provide the source \LaTeX{} string for that reason. Another possible improvement in recognition can come from additional contextual information, for instance the scientific field \cite{Collier2022TransferAM} that can be added post-hoc. Combining recognizers with a language model \cite{carbune2020fast} trained on a large set of mathematical expressions would be a step in a similar direction.

\section{Limitations}\label{sec:limits}
A single sample in MathWriting dataset has one handwritten \LaTeX{} formula, see Figure~\ref{fig:ink-example}. As a result, models that are trained on this dataset would perform poorly on complete handwritten documents, such as the IAMonDo dataset \cite{IAMonDo}. Also, as the dataset contains only \LaTeX{} expressions, it is unlikely that models trained on it will accurately recognize handwritten text in English or other languages. As shown in Figure~\ref{fig:top_100_tokens}, some \LaTeX{} tokens are way more frequent than others. Some infrequent tokens like \verb|\ni| could be hard to recognise.

\section{Conclusion}
We introduced MathWriting, the largest dataset of online handwritten mathematical expressions to date, together with the experimental results of three different types of models. We hope this dataset will help advance research in both online and offline mathematical expression recognition. Additionally, we invite data practitioners to build on the dataset. We intentionally chose a file format for MathWriting close to the one used by CROHME to facilitate their combined use. We also provided original or intermediate representations (raw \LaTeX{} strings, bounding boxes) to enable experimentation with the data itself, and suggested a few directions (Section~\ref{ref:applications_future}).

\subsection*{Acknowledgements}

We warmly thank all the contributors without whom this dataset would not exist. We thank Henry Rowley and Thomas Deselaers for their contribution to organizing the data collection and supporting this effort for many years.  We thank Ashish Myles and MH Johnson for related contributions that resulted in important data and model improvements. We thank Vojta Letal and Pedro Gonnet for their contributions to the CTC Transformer model as well as to synthetic data. We thank Peter Garst and Jonathan Baccash for their contribution to the label normalizer. We thank Blagoj Mitrevski and Henry Rowley for their useful suggestions regarding the text of the paper. We thank our product counsels Janel Thamkul and Rachel Stigler for their legal advice.


\printbibliography

\appendix
\section*{Appendix}

\section{\LaTeX\ template for label rendering}\label{appendix:latex_template}

All the packages and definitions that are required to compile all the normalized and raw labels:

\begin{verbatim}
\usepackage{amsmath}
\usepackage{amsfonts}
\usepackage{amssymb}
\newcommand{\R}{\mathbb{R}}
\newcommand{\C}{\mathbb{C}}
\newcommand{\Q}{\mathbb{Q}}
\newcommand{\Z}{\mathbb{Z}}
\newcommand{\N}{\mathbb{N}}
\end{verbatim}

\section{Acquisition of \LaTeX{} Expressions}\label{sec:wikipedia_labels}
The labels we publish mostly come from Wikipedia (95\% of all samples have labels from Wikipedia). A small part were generated, to cover deeply nested fractions, number-heavy expressions, and isolated letters with nested superscripts and subscripts, which are rare in Wikipedia.

The extraction process from Wikipedia followed these steps:

\begin{itemize}
    \item download an XML Wikipedia dump which provides Wikipedia's raw textual content. \texttt{enwiki-20231101-pages-articles.xml} was used for synthetic samples, older dumps for human-written ones
    \item extract all \LaTeX{} expressions from that file. This gives the list of all mathematical expressions in \LaTeX{} notation from Wikipedia
    \item keep those which could be compiled using the packages listed in Appendix~\ref{appendix:latex_template}. Wikipedia contains a significant number of expressions that are not accepted by the \LaTeX{} compiler, because of syntax errors or other reasons
    \item keep only those which can be processed by our normalizer which only supports a subset of all \LaTeX{} commands and structures
\end{itemize}

For expressions used for synthesis, the following extra steps were performed:

\begin{itemize}
    \item keep only the expressions whose normalized form contains more than a single \LaTeX{} token. Example: \verb|\alpha| is rejected but \verb|\alpha^{2}| is kept. This step is useful to eliminate trivial expressions that wouldn't add any useful information
    \item de-duplicate expressions based on their normalized form. e.g. \verb|\frac12| and \verb|\frac{1}{2}| normalize to the same thing, we kept only one of them in raw form
    \item restrict the list of expressions to the same set of tokens used in the \train{} split: if the normalized form of an expression contained at least one token that was not also present somewhere in \train{}, it was discarded.
\end{itemize}


\section{Postprocessing of MathWriting dataset}\label{sec:postprocessing}

We applied no postprocessing to the collected inks other than dropping entirely those that were completely unreadable or had stray marks. Inks are provided in their original form, as they were recorded with the collection app. What we \textit{did not do} was to discard samples that were very hard to read or ambiguous, because we believe this type of sample to be essential in training a high-quality model.

Some cleanup was performed on the labels (ground truths). The goal was to make the dataset better suited to training an ML model, and eliminate unavoidable issues that occurred during the collection. After training some initial models, we manually reviewed samples for which they performed poorly. This helped identify a lot of unusable inks (near-blank, lots of stray strokes, scribbles, etc.), and a lot of ink/label discrepancies. A fairly common occurrence was a contributor forgetting to copy part of the prompted expression. We adjusted the label to what was actually written unless the ink contained a partially-drawn symbol, in which case we discarded the sample entirely. In this process we eliminated or fixed around 20k samples.

The most important postprocessing step was to normalize the labels: there are many different ways to write a mathematical expression in \LaTeX{} format that will render to images that are equivalent in handwritten form. We applied a series of transformations to eliminate as many variations as possible while retaining the same semantic. This greatly improved the performance of models and made their evaluation more precise. We publish both the normalized and raw (unnormalized) labels, to enable people to experiment with other normalization procedures. 

This normalization is similar to what~\cite{deng2017imagetomarkup} did, but pushed further because of the specifics of handwritten MEs. See Section~\ref{sec:normalization} for more detail.
\section{Dataset split}\label{sec:dataset_split}
The \valid{} and \test{} splits are the result of multiple operations performed between 2016 and 2019. The first split operation, performed on the data available in 2016, was based on the contributor id: any given contributor's samples would not appear in more than one split (either \train, \valid, \test). This is common practice for handwriting recognition systems, to test how the recognizer performs on unseen handwriting styles.

Experiments then showed that a more important factor than the handwriting style was whether the \textit{label} had already been seen during training. Subsequent data collection campaigns focused on increasing label variety, and new samples were added to \valid{} and \test, this time split by label: a given normalized mathematical expression would not appear in more than one split.

\section{Label Normalization}\label{sec:appendix_normalization}

\begin{figure*}
\begin{center}

\begin{tabular}{ll}
\hline\\[-2pt]
\multicolumn{2}{l}{Sub/superscript are put in braces, \texttt{\textbackslash over} is replaced by \texttt{\textbackslash frac}} \\[4pt]
Raw: & \verb|\overline{hu^2}+{1 \over 2}{k_{ap}g_zh^2}| \\
Normalized: & \verb|\overline{hu^{2}}+\frac{1}{2}k_{ap}g_{z}h^{2}| 
\\[6pt] \hline\\[-2pt]

\multicolumn{2}{l}{Subscripts are put before superscripts, extra space is dropped} \\[4pt]
Raw: & \verb|\int^a_{-a}f(x) dx=0| \\
Normalized: & \verb|\int_{-a}^{a}f(x)dx=0| 
\\[6pt] \hline\\[-2pt]

\multicolumn{2}{l}{Single quotes are replaced by a superscript} \\[4pt]
Raw: & \verb|f'(\overline x)| \\
Normalized: & \verb|f^{\prime}(\overline{x})| 
\\[6pt] \hline\\[-2pt]

\multicolumn{2}{l}{Text formatting commands like \texttt{\textbackslash rm} are dropped} \\[4pt]
Raw: & \verb|~A_{0}=\frac{ND}{\sigma_{\rm as}+\sigma_{\rm es}}~| \\
Normalized: & \verb|A_{0}=\frac{ND}{\sigma_{as}+\sigma_{es}}| 
\\[6pt] \hline\\[-2pt]

\multicolumn{2}{l}{\parbox{14cm}{Matrix environments with delimiters like \texttt{bmatrix} are replaced by \texttt{matrix} surrounded by delimiters\\ Commands like \texttt{\textbackslash cos} are replaced by the series of letters}} \\[11pt]
Raw: & \verb|\begin{bmatrix} -\sin t \\ \cos t \end{bmatrix}| \\
Normalized: & \verb|[\begin{matrix}-sint\\ cost\end{matrix}]| 
\\[6pt] \hline\\[-2pt]

\multicolumn{2}{l}{Delimiter size modifiers like \texttt{\textbackslash big} are dropped} \\[4pt]
Raw: & \verb|\big(\tfrac{a}{N}\big)| \\
Normalized: & \verb|(\frac{a}{N})|
\\[6pt] \hline\\[-2pt]

\end{tabular}
\end{center}
\caption{Examples of expression normalization. See Section \ref{sec:normalization} for details.}
\label{fig:normalized-labels}
\end{figure*}

\subsection{Syntactic Variations}

There are several ways to change a \LaTeX{} string without changing the rendered output significantly. The normalization we implemented does the following:

\begin{itemize}
    \item all unnecessary space is dropped
    \item all command arguments are consistently put in curly braces
    \item superscripts and subscripts are put in curly braces and their order is normalized. e.g. \verb|a^2_1| becomes \verb|a_{1}^{2}|.
    \item redundant braces are dropped
    \item infix commands are replaced by their prefix versions. e.g. \verb"\over" is replaced by \verb"\frac"
    \item a lot of synonyms are collapsed. e.g. \verb"\le" and \verb"\leq", 
    \verb"\longrightarrow" and \verb"\rightarrow", etc. Some of the synonyms are only synonyms in handwriting. For example \verb"\star" ($\star$) and $*$ are different in print (5-prong and 6-prong stars), but the difference was not expressed in handwriting by our contributors.
    \item functions commands like \verb"\sin" are replaced by the sequence of letters of the function name (e.g. \verb"\sin" is replaced by \verb"sin"). This reduces the output vocabulary, and eliminates a source of confusion because we found that \LaTeX{} expressions from Wikipedia come with a mix of function commands and sequences of letters.
    \item expansion of abbreviations. e.g. \verb"\cdots", \verb"\ldots", etc. have been replaced by the corresponding sequence of characters.
    \item matrix environments are normalized to use only the 'matrix' environment surrounded by the proper delimiters like brackets or parentheses.
    \item \verb|\binom| is turned into a 2-element column matrix. Expressions from Wikipedia did not use those consistently, so we made the choice to normalize \verb|\binom| away.
\end{itemize}

\subsection{Differences Between Print And Handwriting}

The following characteristics can not be represented in handwriting and have been normalized away:

\begin{itemize}
\item color
\item accurate spacing: e.g. \verb|~|, \verb|\quad|.
\item font style and size: e.g. \verb|\mathrm|, \verb|\mathit|, \verb|\mathbf|, \verb|\scriptstyle|.
\end{itemize}

There are others that can be represented in handwriting, but that are not consistent enough in MathWriting to be preserved:

\begin{itemize}
\item font families: Fraktur, Calligraphic. In practice, only Blackboard (\verb"\mathbb") has been written consistently enough by contributors that we were able to keep it: \verb"\mathcal" and \verb"\mathfrak" are dropped.
\item some variations like \verb"\rightarrow" $\rightarrow$ and   \verb"\longrightarrow" $\longrightarrow$.
\item some character variations. e.g. \verb|\varrho|, \verb|\varepsilon|
\item size modifiers like \verb"\left", \verb"\right", \verb"\big". Similarly, variable-width diacritics like \verb"\widehat".
\end{itemize}

\section{Additional dataset statistics}\label{sec:dataset_additional}

\begin{figure*}[!ht]
    \centering
    \includegraphics[width=\textwidth]{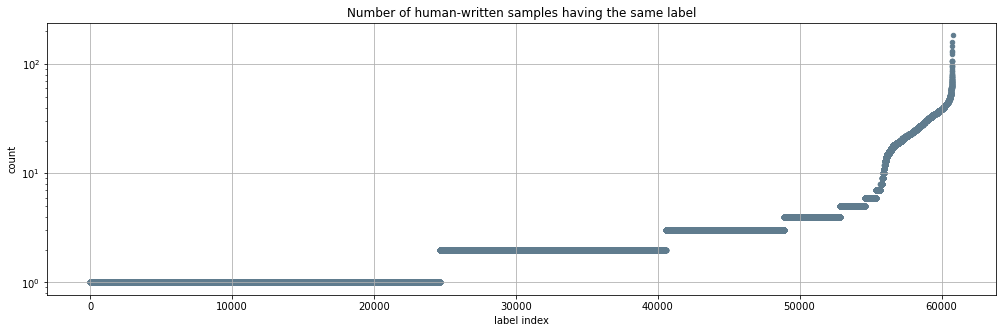}
    \cprotect\caption{Counts of inks corresponding to the same normalized expression, ordered by increasing count. Each position on the horizontal corresponds to a unique normalized expression. Almost 5k unique expressions have been written 10 times or more by contributors.
    }
    \label{fig:frequency_plot}
\end{figure*}

\begin{figure*}[!ht]
    \centering
    \includegraphics[width=\textwidth]{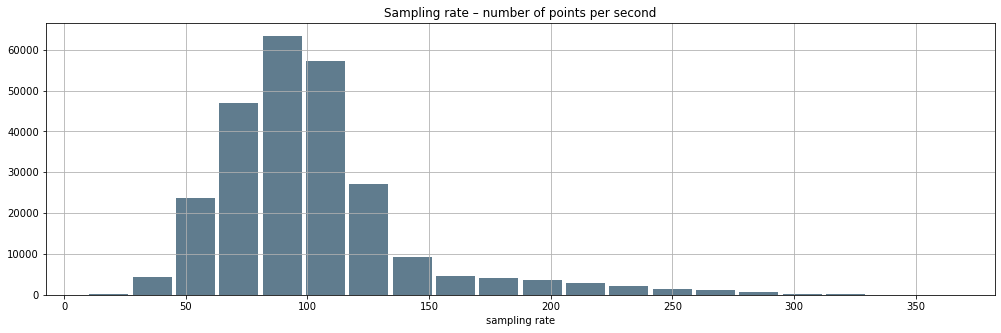}
    \cprotect\caption{Histogram of sampling rates in human-written data of MathWriting dataset.}
    \label{fig:sampling_rate}
\end{figure*}

In this section we show additional graphs that illustrate dataset statistics that are described in Section~\ref{sec:data_stat}. The frequencies of normalized \LaTeX{} expressions are presented in Figure~\ref{fig:frequency_plot}. Figure~\ref{fig:sampling_rate} illustrates the distribution of sampling rates within human-written data. Results of resampling points in time are presented in Figure~\ref{fig:time_sampling}.


\begin{table}
\begin{center}
\begin{tabular}{lr}
\toprule
Device type & Ink \\ \midrule
Google PixelBook   & 51k \\
Google Nexus 5X    & 28k \\
Coolpad Mega 2.5D  & 14k \\
OnePlus One        & 13k \\
Google Nexus 5     & 11k \\
Google Nexus 6     & 11k \\
Google Nexus 6P    & 11k \\
Coolpad Mega 3     & 8k \\
LG Optimus L9      & 8k \\ 
Galaxy Grand Duos  & 7k \\
Google Pixel XL    & 6k \\
Samsung Galaxy S7  & 5k \\
\bottomrule
\end{tabular}
\end{center}
\caption{Top-12 devices used, with the number of samples obtained from each device. The bias towards Google devices simply reflects the conditions in which inks were collected.}
\label{tab:devices}
\end{table}
\section{Variety of Writing Styles}\label{sec:discussion_additional}
\begin{figure}
    \begin{center}
    \inkimg{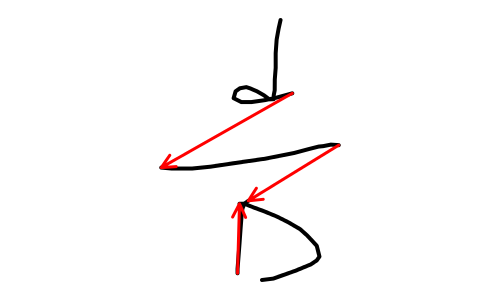}{a751880b939d5a9a}{130pt}
    \inkimg{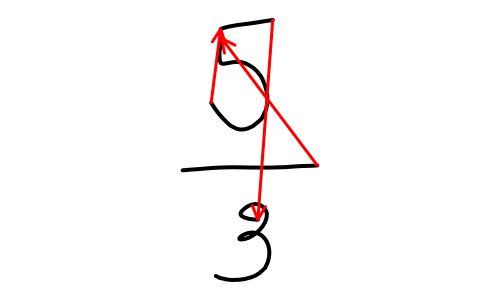}{8c95114b04a97aa2}{130pt}\\
    \vspace{1em}
    \inkimg{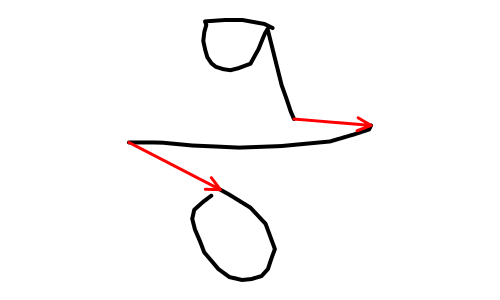}{0348238e894e8d62}{130pt}
    \inkimg{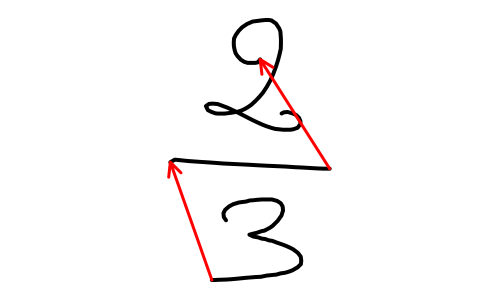}{ee557c63b5755a6f}{130pt}
    \end{center}
    \caption{Examples of various writing orders found in the training set. Red arrows show the movement of the pen between strokes.
    The top one: most common writing order (top-down, fraction bar drawn left-to-right), second from the top: fraction bar written first, third from the top: fraction bar drawn right-to-left, forth from the top: fraction written bottom-up.}
    \label{fig:writingstyle_fractions}
\end{figure}

In this section we provide additional examples of differences in the writing order of fractions – Figure~\ref{fig:writingstyle_fractions}. These examples show that MathWriting dataset contains a variety of writing styles.


\section{Sources of Noise}\label{sec:source_noise}
The result of any task performed by humans will contain mistakes, and MathWriting is no exception. We've done our best to remove most of the mistakes, but we know that some remain.

\paragraph{Stray strokes} These do not carry any meaning and should be ignored by any recognizer. Since they also appear in real applications, there could be some benefit in having some in the dataset to teach the model about them. That said, it being usually easier to add noise rather than to remove it, we made the choice of discarding as many inks containing stray strokes as possible. Not all inks with stray strokes have been found and removed though (e.g. \verb|train/9e64be65cb874902.inkml| that was discovered post-publication). The fraction of inks containing stray strokes is significantly lower than 1\%, and should not be an issue for training a model.

\paragraph{Incorrect ground truth} Contributors did not always copy the prompt perfectly, leading to a variety of differences. In most of the cases we spotted, we were able to fix the label to match what had actually been written. A short manual review once the dataset was in its final state showed the rate of incorrect ground truth to be between 1\% and 2\%. Most of the mistakes are very minor, usually a single token added, missing or incorrect. Errors here also come from ambiguities or misuse of the \LaTeX{} notation: expressions coming from Wikipedia contain some misuse like using \verb|\Sigma| where \verb|\sum| was more appropriate, \verb|\triangle| instead of \verb|\Delta|, \verb|\triangledown| instead of \verb|\nabla|, \verb|\begin{matrix}\end{matrix}| instead of \verb|\binom|, and also some handwriting-specific ambiguities like \verb|\dagger| vs \verb|\top| vs \verb|T|. There are also some instances where reference numbers or extra punctuation are included.


\paragraph{Aggressive normalization} While the above sources of noise are unavoidable, normalization is a postprocessing operation that can in theory be tweaked to perfection. In practice, it's a compromise between reducing accidental ambiguities (i.e. removing synonyms), and removing information. Examples: we made the choice of treating \verb|\binom| as a synonym for a 2-element matrix. While it does improve recognition accuracy by making the problem easier, it also moves the burden of distinguishing between the two cases to downstream steps in the recognition pipeline. Similar things can be said about removing all commands that indicate that their content is text instead of math (e.g. \verb|\mbox|), dropping size modifiers, rewriting function commands (e.g. \verb|\sin|, \verb|\cos|), etc. Using a different normalization could prove beneficial depending on the context the recognizer is used in practice. However, for the purpose of a benchmark any reasonable compromise is adequate.

\section{Tokens}\label{sec:tokens}


Using the above code to compute tokens, the set of all samples in the dataset (human-written, synthetic, from all splits) contain the following after normalization:

\begin{itemize}

\item Syntactic tokens: \_ \textasciicircum \{ \} \& \textbackslash\textbackslash\ space

\item Latin letters and numbers: a-z A-Z 0-9
\item Blackboard capital letters \verb|\mathbb{A}|-\verb|\mathbb{Z}| \verb|\mathbb|

\item Latin punctuation and symbols: , ; : ! ? . ( ) [ ] \verb|\{| \verb|\}| 
 * / + - \verb|\_| \verb|\&| \verb|\#| \verb|\%| $|$ \verb|\backslash|

\item Greek letters:
\verb|\alpha| \verb|\beta| \verb|\delta| \verb|\Delta| \verb|\epsilon| \verb|\eta| \verb|\chi| \verb|\gamma| \verb|\Gamma| \verb|\iota| \verb|\kappa| \verb|\lambda| \verb|\Lambda| \verb|\nu| \verb|\mu| \verb|\omega| \verb|\Omega| \verb|\phi| \verb|\Phi| \verb|\pi| \verb|\Pi| \verb|\psi| \verb|\Psi| \verb|\rho| \verb|\sigma| \verb|\Sigma| \verb|\tau| \verb|\theta| \verb|\Theta| \verb|\upsilon| \verb|\Upsilon| \verb|\varphi| \verb|\varpi| \verb|\varsigma| \verb|\vartheta| \verb|\xi| \verb|\Xi| \verb|\zeta|

\item Mathematical constructs:
\verb|\frac| \verb|\sqrt| \verb|\prod| \verb|\sum| \verb|\iint| \verb|\int| \verb|\oint| 

\item Diacritics and modifiers - 
Note the absence of the single-quote character, which is normalized to \verb|^{\prime}|: \\
\verb|\hat| \verb|\tilde| \verb|\vec| \verb|\overline| \verb|\underline| \verb|\prime|
\verb|\dot| \verb|\not|

\item Matrix environment: \verb|\begin{matrix}| \verb|\end{matrix}|

\item Delimiters: 
\verb|\langle| \verb|\rangle| \verb|\lceil| \verb|\rceil| \verb|\lfloor| \verb|\rfloor| \verb"\|"

\item Comparisons:
\verb|\ge| \verb|\gg| \verb|\le| \verb|\ll| \textless \textgreater

\item Equality, approximations:
\verb|=| \verb|\approx| \verb|\cong| \verb|\equiv| \verb|\ne| \verb|\propto| \verb|\sim| \verb|\simeq| 

\item Set theory:
\verb|\in| \verb|\ni| \verb|\notin| \verb|\sqsubseteq| \verb|\subset| \verb|\subseteq| \verb|\subsetneq| \verb|\supset| \verb|\supseteq| \verb|\emptyset|

\item Operators:
\verb|\times| \verb|\bigcap| \verb|\bigcirc| \verb|\bigcup| \verb|\bigoplus| \verb|\bigvee| \verb|\bigwedge| \verb|\cap| \verb|\cup| \verb|\div| \verb|\mp| \verb|\odot| \verb|\ominus| \verb|\oplus| \verb|\otimes| \verb|\pm| \verb|\vee| \verb|\wedge|

\item Arrows: \verb|\hookrightarrow| \verb|\leftarrow| \verb|\leftrightarrow| \verb|\Leftrightarrow| \verb|\longrightarrow| \verb|\mapsto| \verb|\rightarrow| \verb|\Rightarrow| \verb|\rightleftharpoons| \verb|\iff|

\item Dots:
\verb|\bullet| \verb|\cdot| \verb|\circ|

\item Other symbols:
\verb|\aleph| \verb|\angle| \verb|\dagger| \verb|\exists| \verb|\forall| \verb|\hbar| \verb|\infty| \verb|\models| \verb|\nabla| \verb|\neg| \verb|\partial| \verb|\perp| \verb|\top| \verb|\triangle| \verb|\triangleleft| \verb|\triangleq| \verb|\vdash| \verb|\Vdash| \verb|\vdots| 

\end{itemize}

\section{Examples of inks}\label{appendix:ink_examples}
This section shows a few examples of rendered inks, so that the reader can get a feel for the kind of data that is in MathWriting. All samples are from the training set. They have been manually picked to show a variety of sizes, characters and structures.

\subsection{Human-Written Samples}
\label{fig:app_samples}

\begin{center}
\inkimg{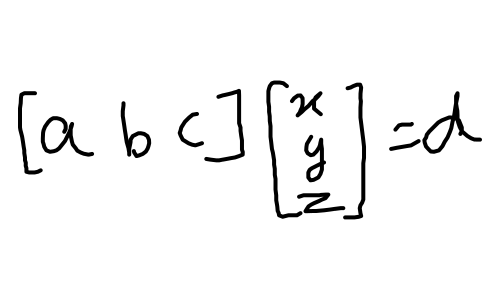}{cf786356546d722c}{100pt}
\inkimg{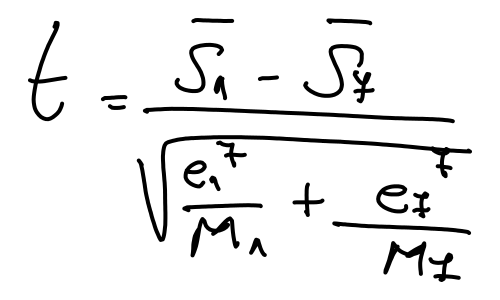}{658e3c257badd8cc}{100pt}
\inkimg{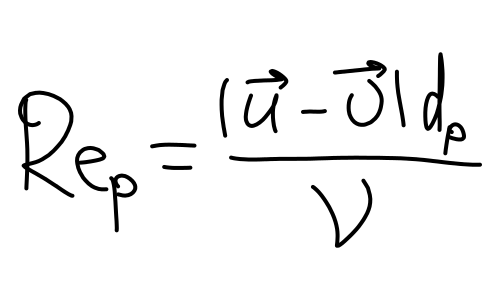}{40b3844b5aeaec00}{100pt}
\inkimg{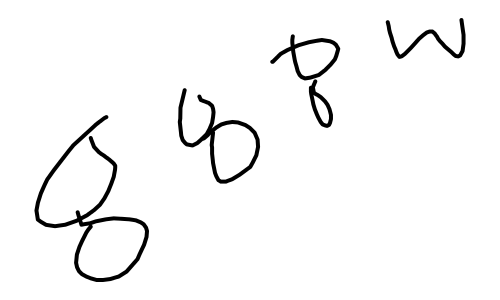}{c28701c7369c22ba}{100pt}
\inkimg{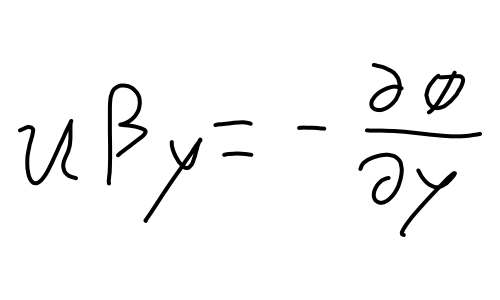}{97e829ac79e851fe}{100pt}
\inkimg{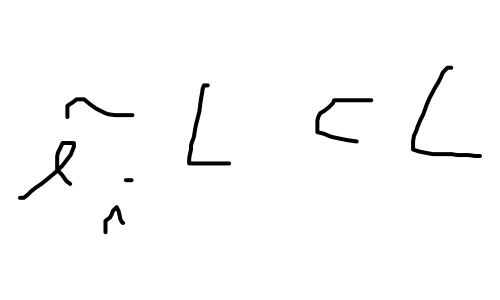}{fd676faba32f1cbb}{100pt}
\inkimg{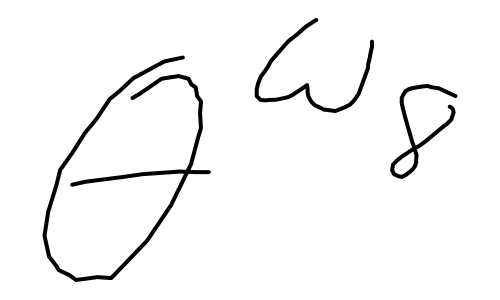}{5e0c81acf5ccdfd3}{100pt}
\inkimg{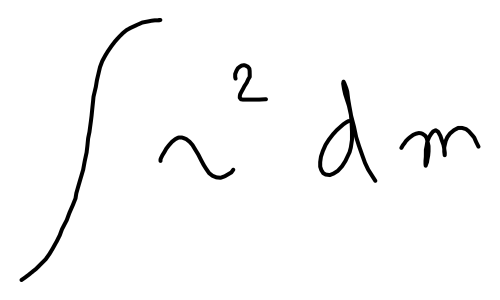}{b3ed172628caafe9}{100pt}
\inkimg{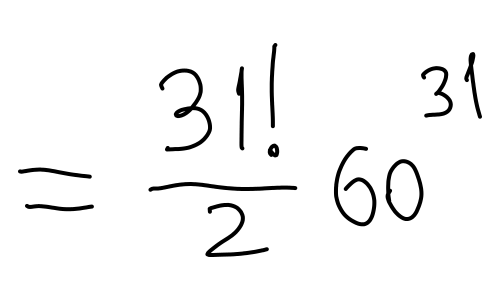}{6150c57b1a98b5ec}{100pt}
\inkimg{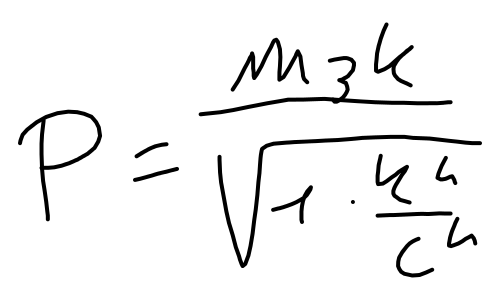}{bca00e3111b70212}{100pt}
\inkimg{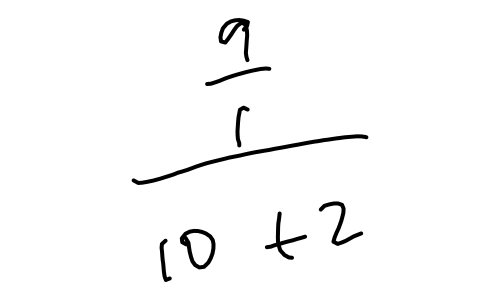}{e829d5eb7e7b3b68}{100pt}
\inkimg{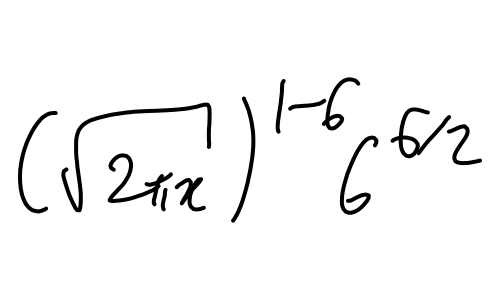}{44bfa5fc08eb5da5}{100pt}
\inkimg{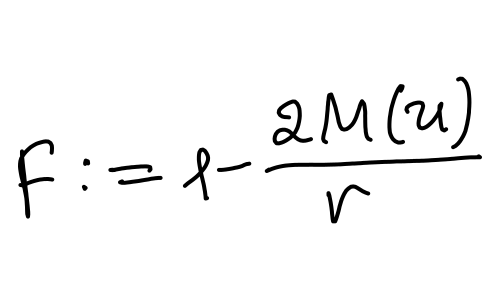}{d233aaf208bd7568}{100pt}
\inkimg{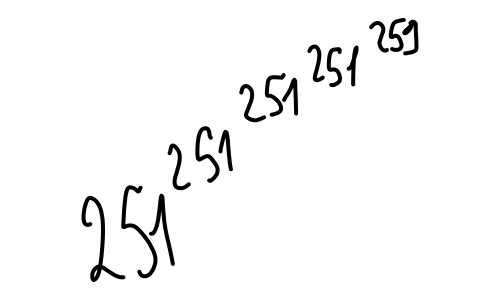}{ee6f0bfd294aa209}{100pt}
\inkimg{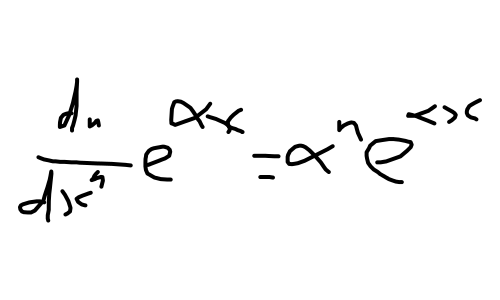}{6579f917f0ba236b}{100pt}
\inkimg{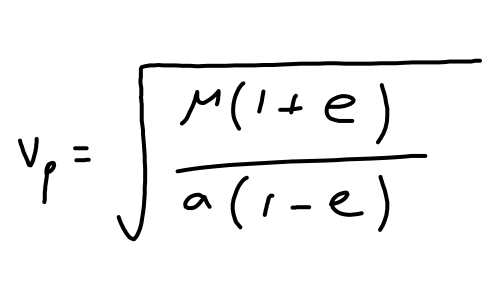}{cea67d239b9f8884}{100pt}
\inkimg{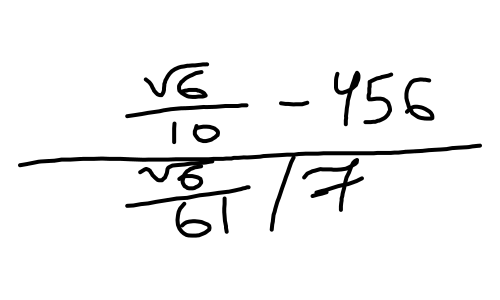}{bb4bca53d0e6336d}{100pt}
\inkimg{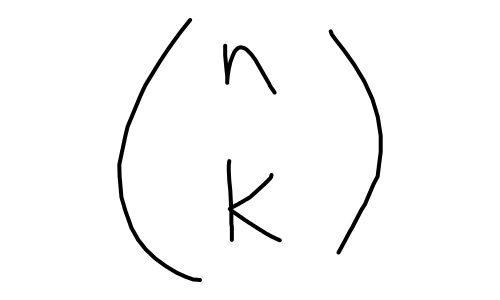}{b14bca3fc2d2819a}{100pt}
\inkimg{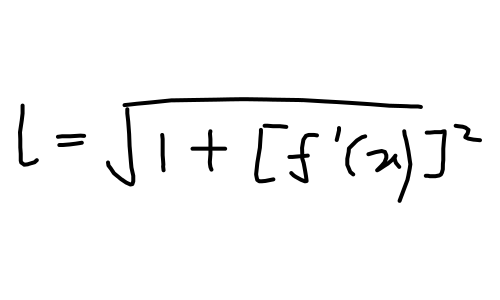}{2409d2feaa79b9d7}{100pt}
\inkimg{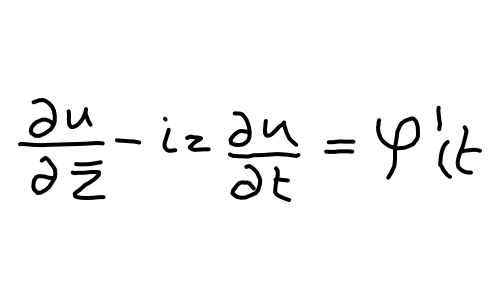}{51486ff88b789d6d}{100pt}
\inkimg{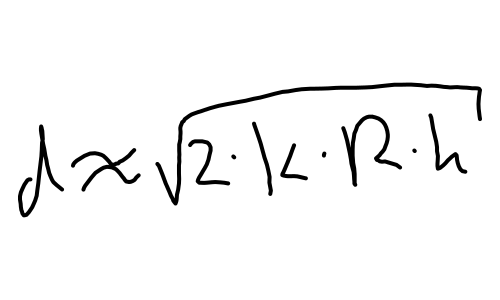}{02229a0c174d8dbe}{100pt}
\inkimg{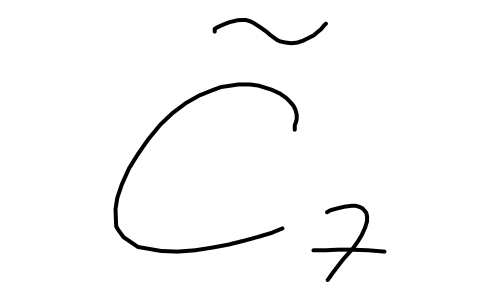}{478e10a15203fa3a}{100pt}
\inkimg{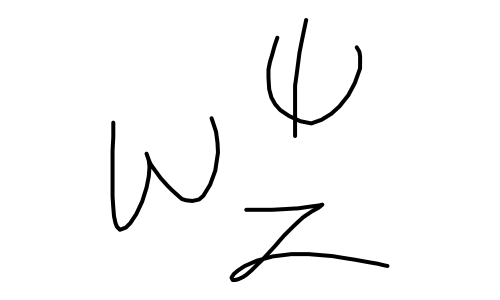}{ccb15825579a096b}{100pt}
\inkimg{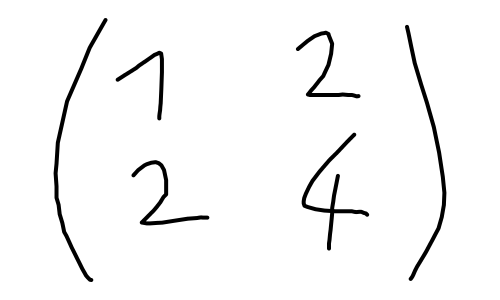}{7a4b95285de0caf0}{100pt}
\inkimg{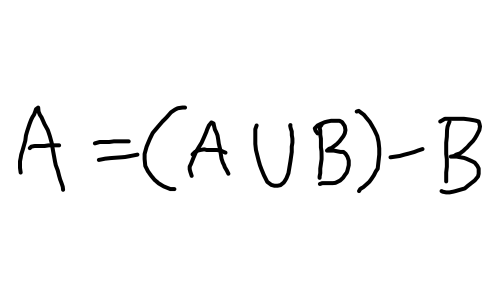}{ac25b4d053596ded}{100pt}
\inkimg{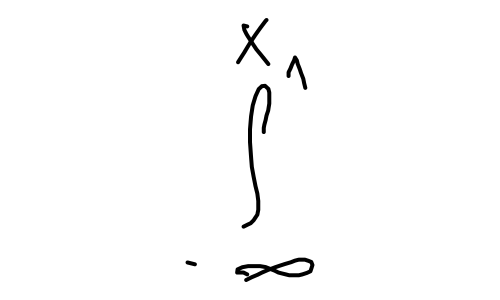}{7d597c52bf8bdd1e}{100pt}
\inkimg{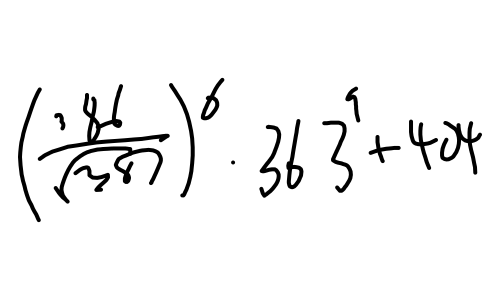}{acc5f6620fad1ce8}{100pt}
\inkimg{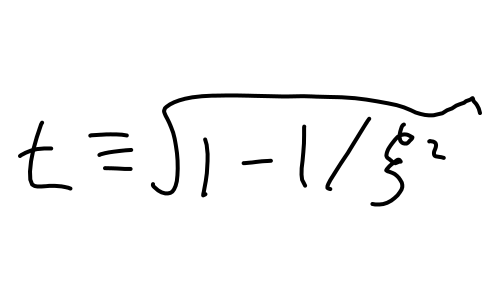}{88c3551d373c72e5}{100pt}
\inkimg{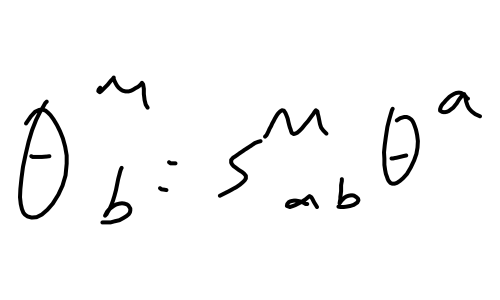}{5009d32d32f80324}{100pt}
\inkimg{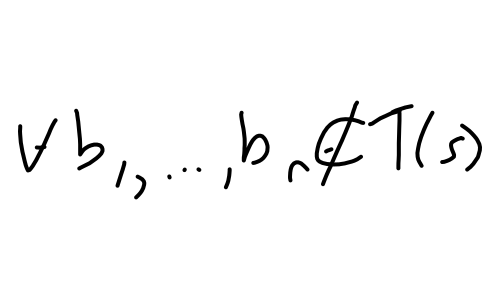}{068de3aad90c403c}{100pt}
\inkimg{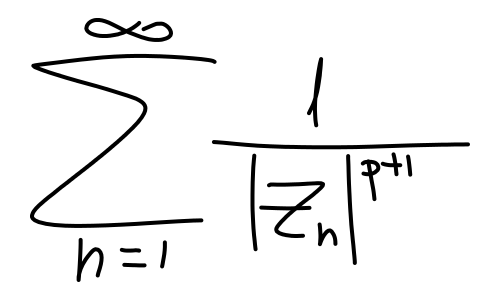}{a3cf115524f0c55b}{100pt}
\inkimg{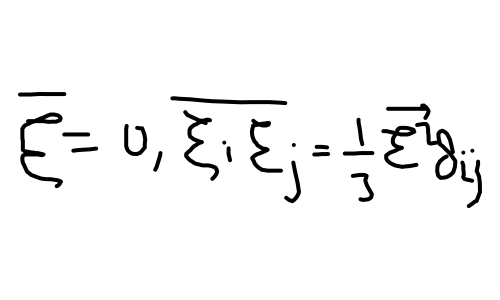}{355f5df56a16913a}{100pt}
\inkimg{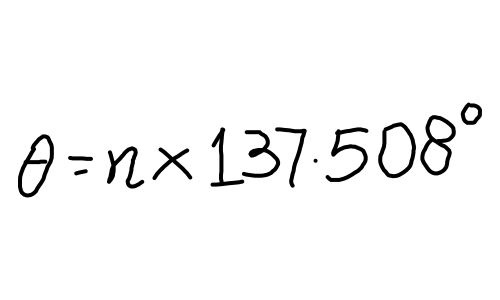}{d9b2ce7aa3495888}{100pt}
\inkimg{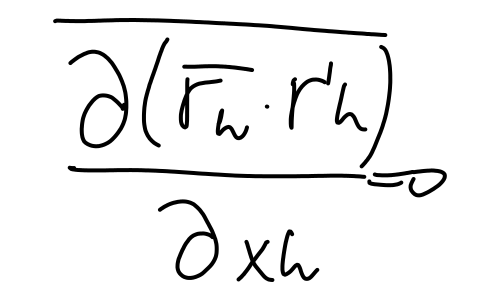}{adceb80fdadf9f6e}{100pt}
\inkimg{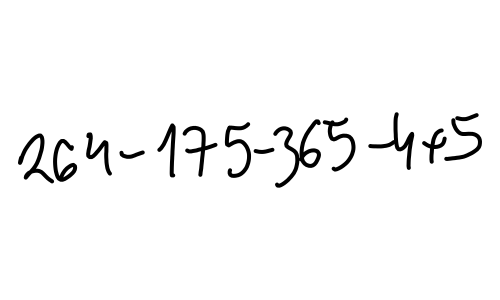}{25892f7caeac8c36}{100pt}
\inkimg{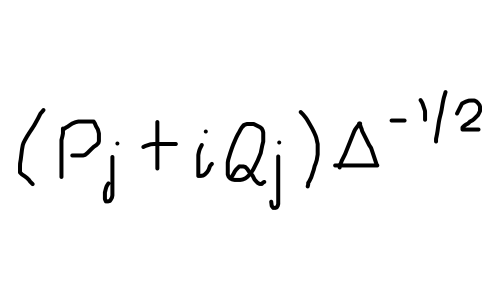}{41e1261a951c6f33}{100pt}
\inkimg{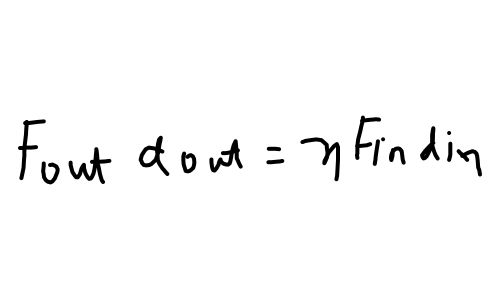}{02a7f7f172671fb4}{100pt}
\inkimg{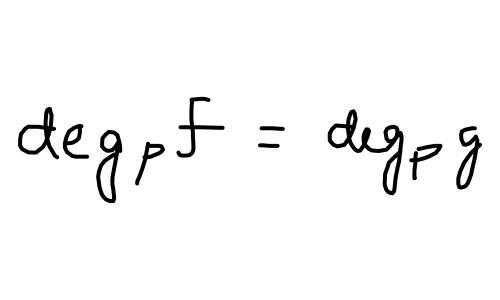}{0d848d4b170d36b9}{100pt}
\inkimg{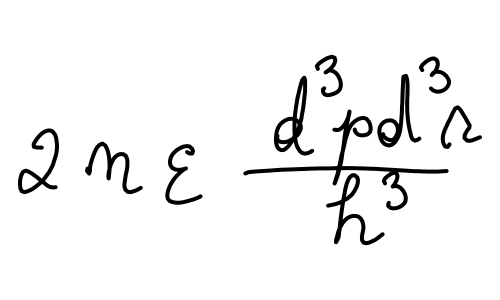}{2adc4f10d42b641c}{100pt}
\inkimg{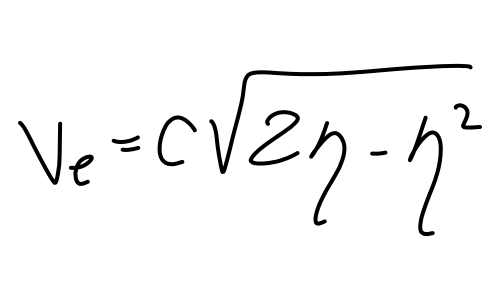}{408f904038dcbba0}{100pt}
\end{center}

\subsection{Synthetic Samples: Expressions from Wikipedia}

\begin{center}
\inkimg{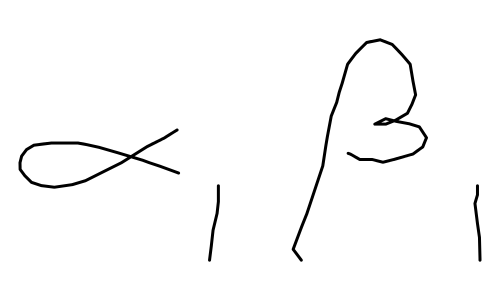}{133829b5a10b783f}{100pt}
\inkimg{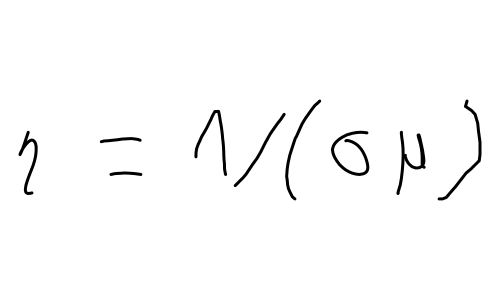}{11623165e9bab0e4}{100pt}
\inkimg{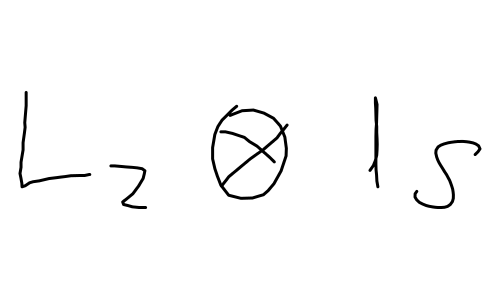}{5ca38d17bf2bea0a}{100pt}
\inkimg{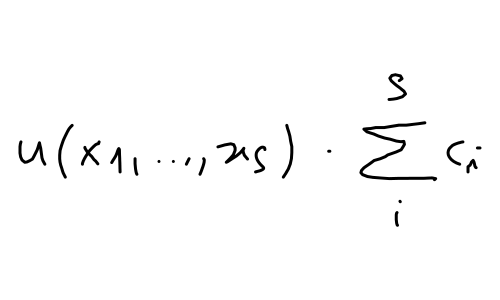}{089650cc894c024b}{100pt}
\inkimg{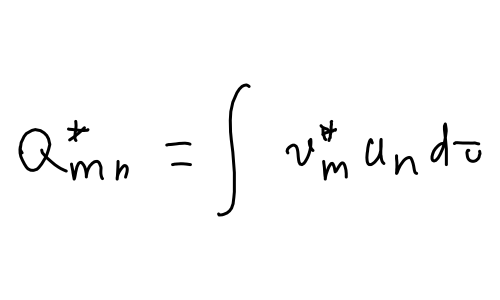}{8e88b75cb5f03bf4}{100pt}
\inkimg{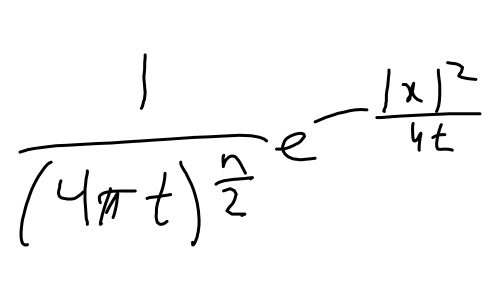}{5c1573b41e762307}{100pt}
\inkimg{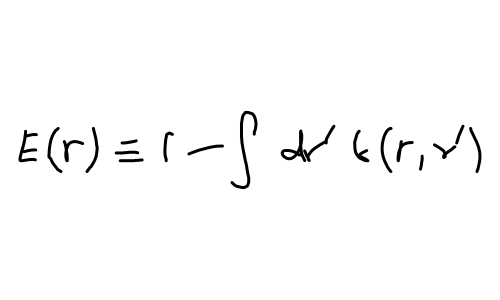}{68e9560a27a093c8}{100pt}
\inkimg{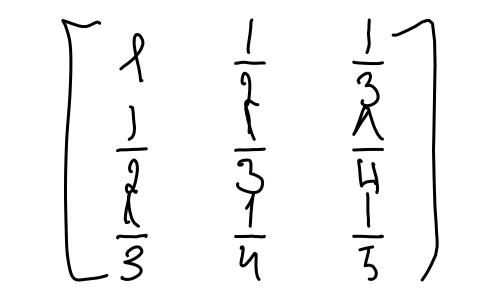}{daab3bae071f8bc0}{100pt}
\inkimg{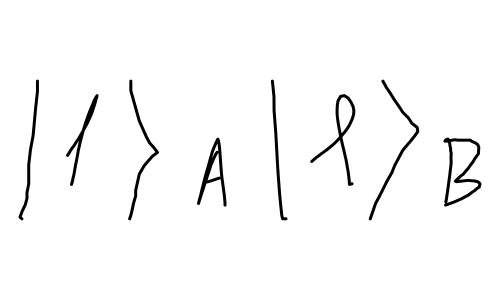}{77602abaea39b774}{100pt}
\inkimg{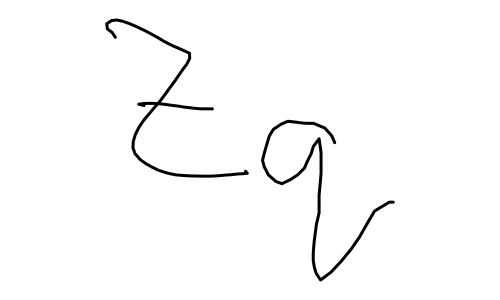}{eb817bcbfd11df18}{100pt}
\inkimg{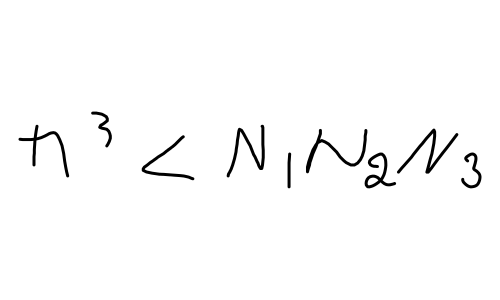}{3bd062b33ea5db6f}{100pt}
\inkimg{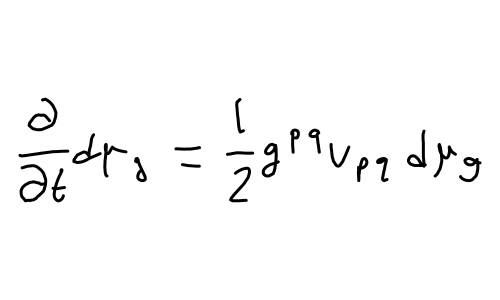}{51ef5122de326151}{100pt}
\inkimg{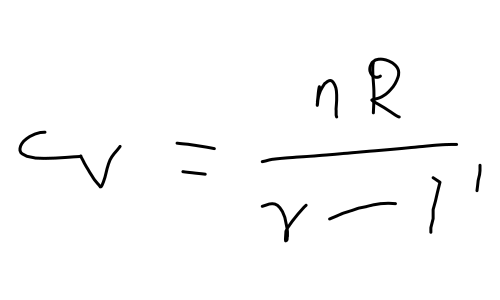}{f2a613fc323df342}{100pt}
\inkimg{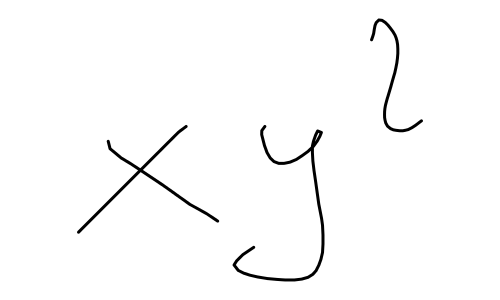}{cbd19abc03ba4098}{100pt}
\inkimg{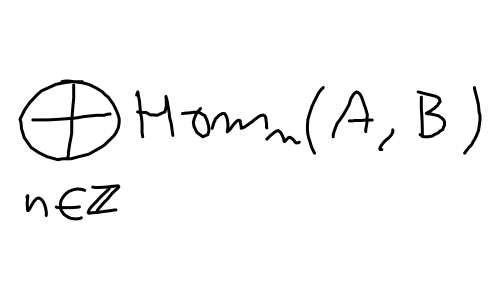}{1879aa5c882b445d}{100pt}
\inkimg{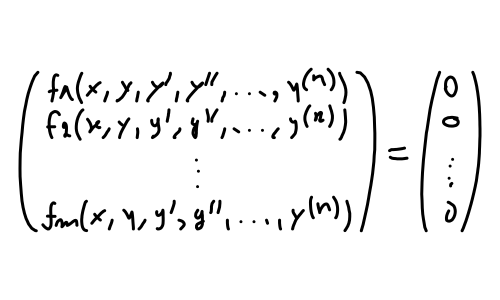}{094c7e52a3f0934d}{100pt}
\end{center}

\begin{center}
\inkimg{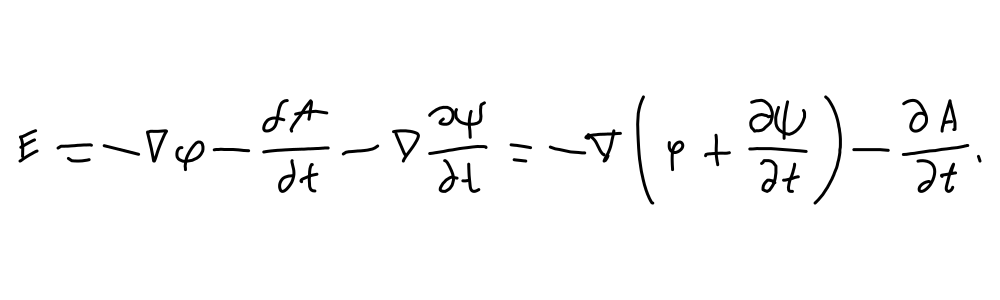}{1cd654228d7ca6bb}{200pt}
\inkimg{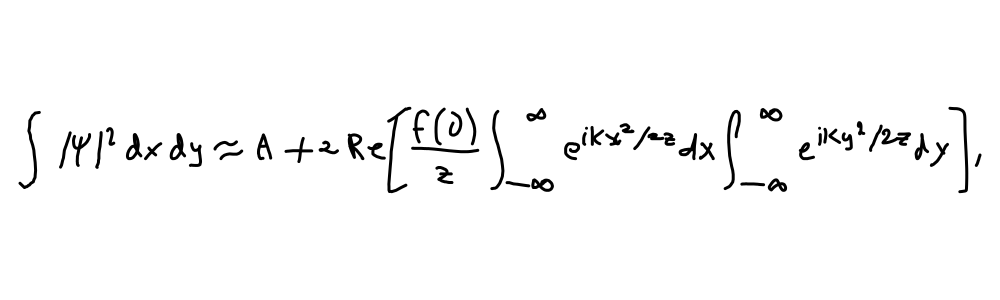}{fc00050933165b70}{200pt}
\inkimg{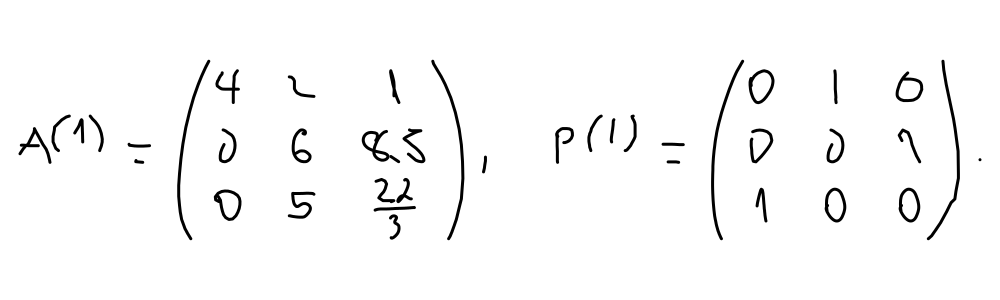}{60553a301aaf84a3}{200pt}
\inkimg{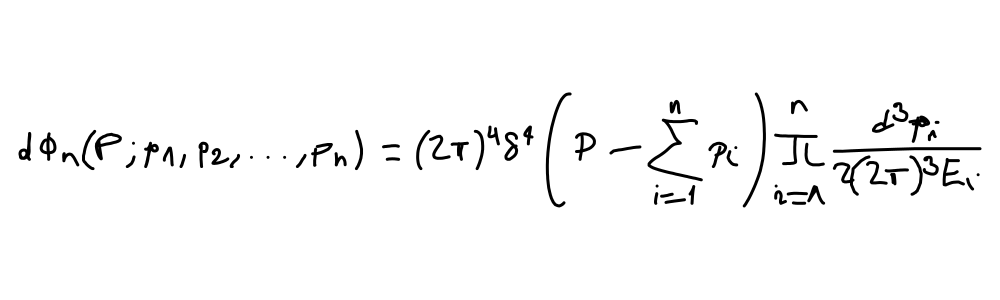}{b2aeaf7a0fd30ed6}{200pt}
\inkimg{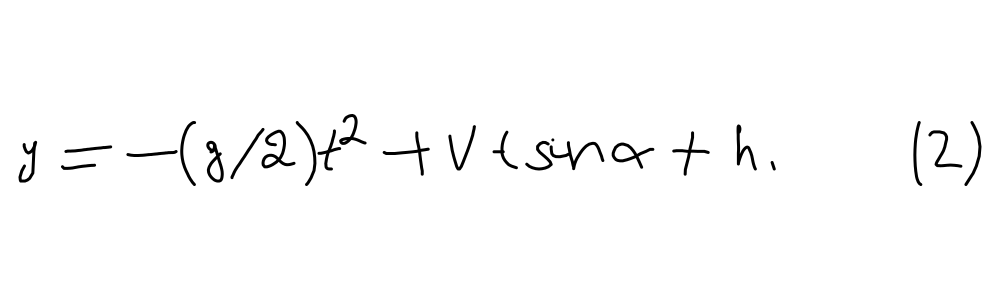}{254dbc2b3843dcf8}{200pt}
\inkimg{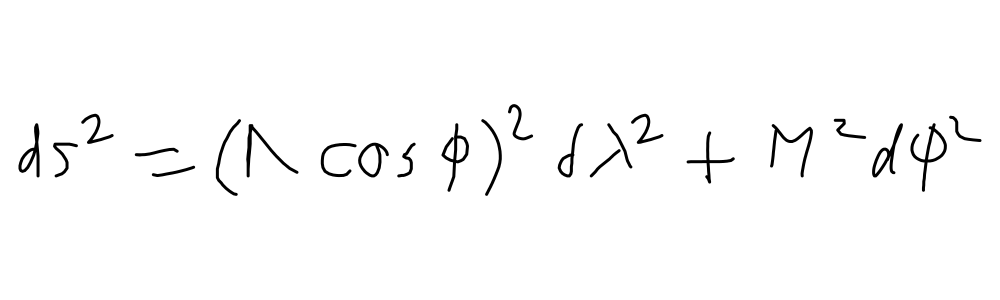}{0772aeaac09d3415}{200pt}
\inkimg{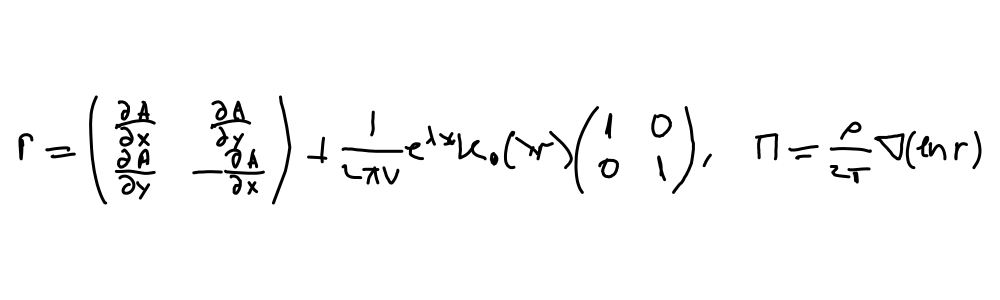}{20b4ebf292cfa8d1}{200pt}
\inkimg{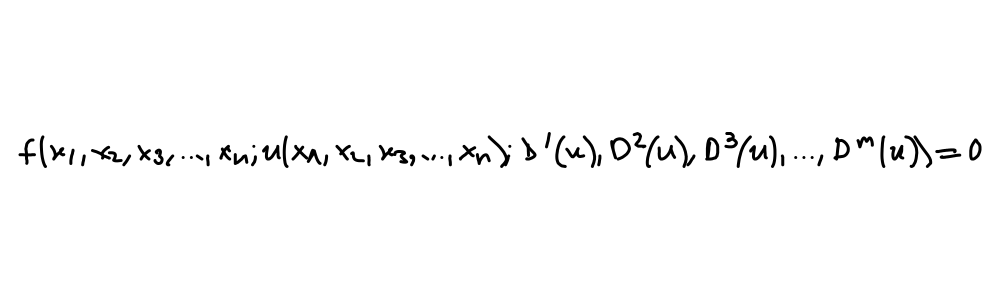}{96613ae167f35f8a}{200pt}
\inkimg{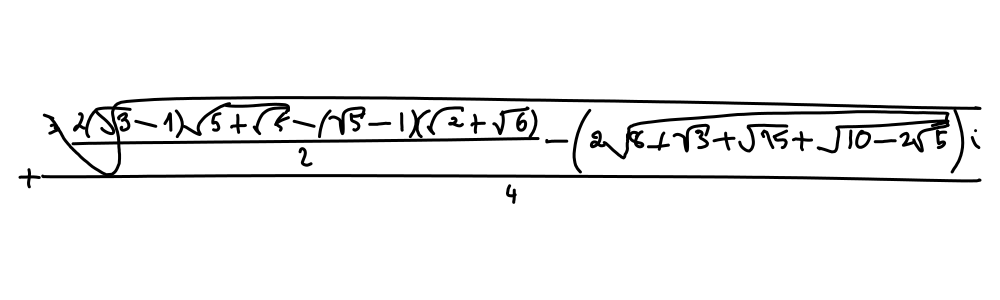}{4b1f5165e3698343}{200pt}
\inkimg{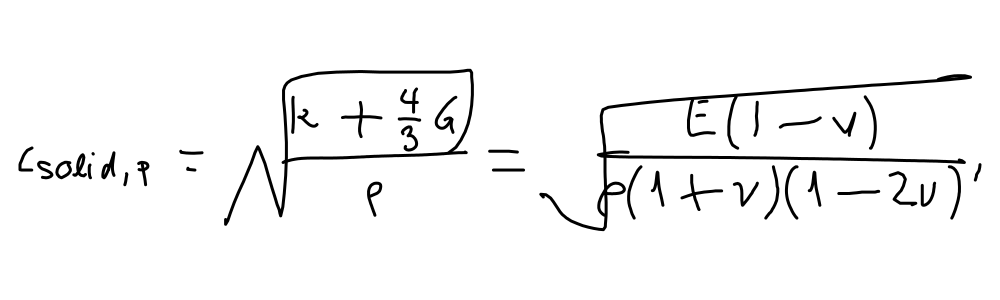}{6f86778996d5f514}{200pt}
\inkimg{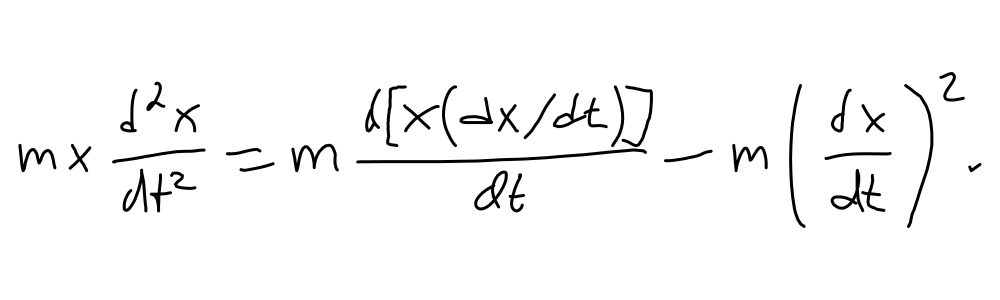}{685bd5676bda74ce}{200pt}
\inkimg{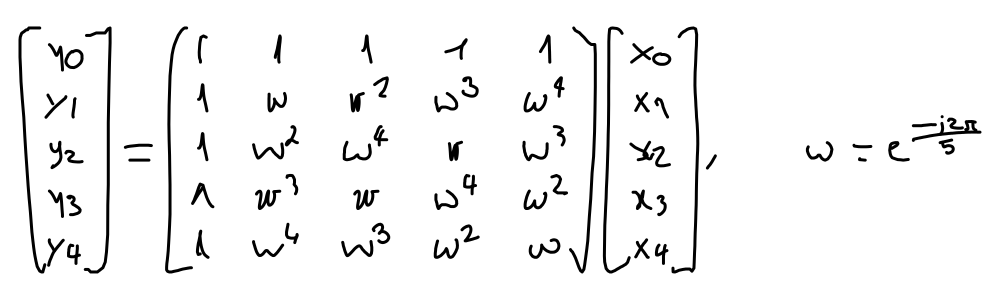}{b79dfccd7f5b5cb4}{200pt}
\end{center}

\subsubsection{Synthetic Samples: Generated Fractions}
\begin{center}
\inkimg{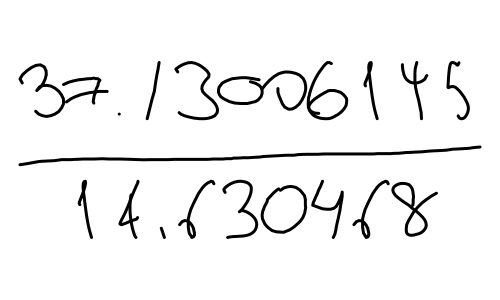}{4677b76acec23465}{100pt}
\inkimg{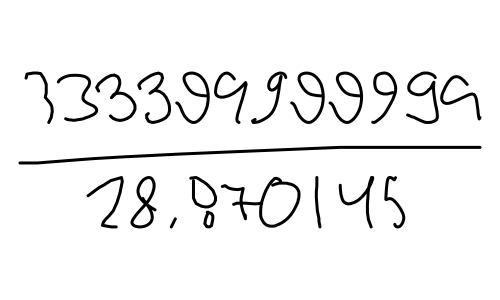}{eef0cd70f8872a9c}{100pt}
\inkimg{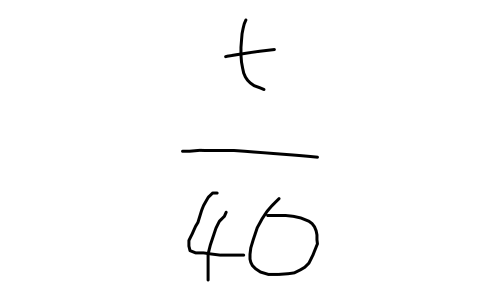}{d9df5fffcfe81d07}{100pt}
\inkimg{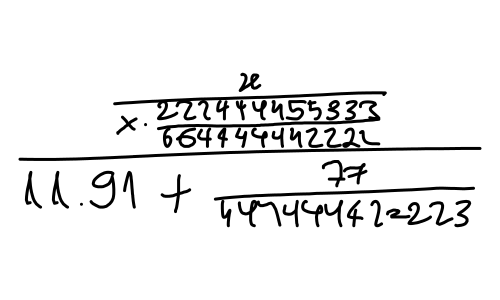}{244ad5e60c9fea92}{100pt}
\inkimg{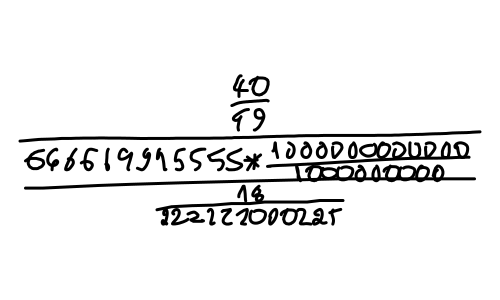}{88d5f862ad46cb47}{100pt}
\inkimg{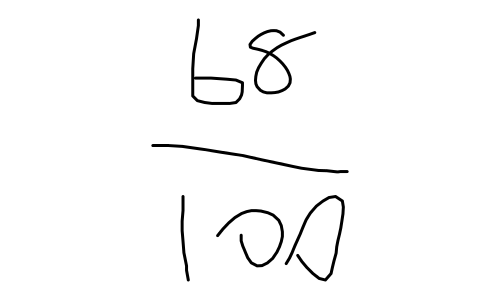}{9067ae238a278b32}{100pt}
\inkimg{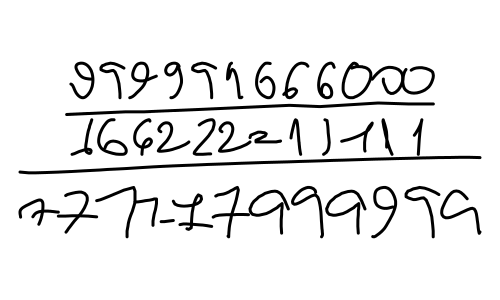}{ce12f955b6ba76a4}{100pt}
\inkimg{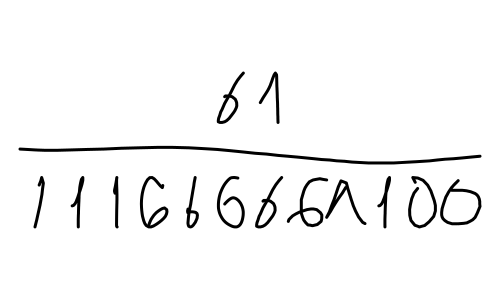}{7fa1aa18a332b211}{100pt}
\inkimg{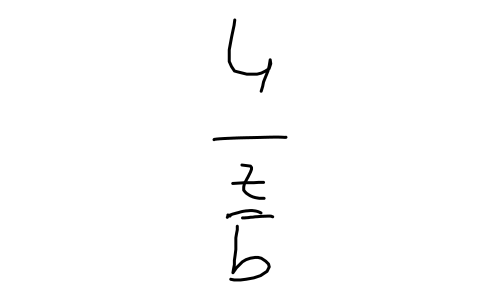}{1c21c51bc1319124}{100pt}
\inkimg{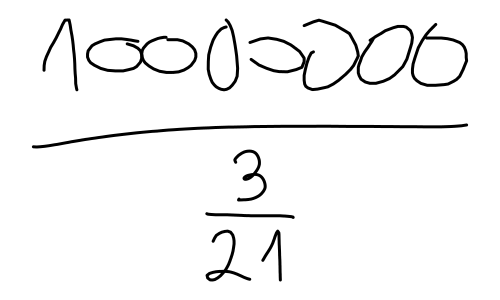}{afd5254b25be1256}{100pt}
\inkimg{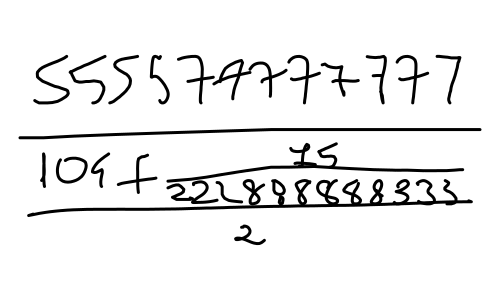}{409a91ba03e3cc7e}{100pt}
\inkimg{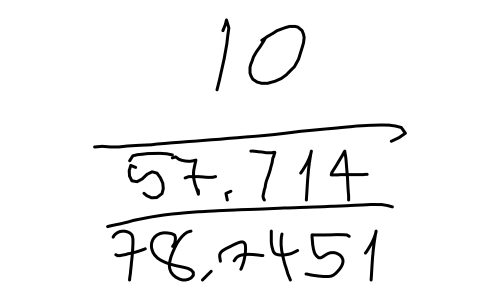}{486a38bd87b8ed97}{100pt}
\inkimg{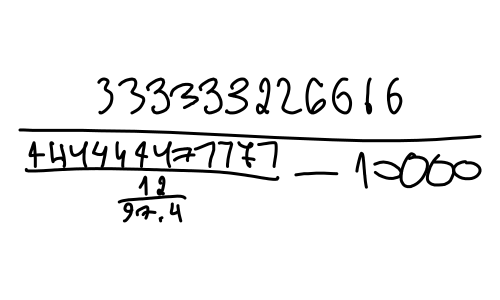}{05efec2565d6726e}{100pt}
\inkimg{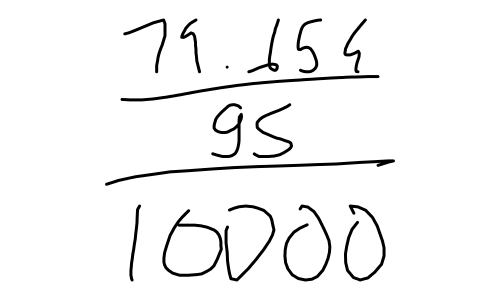}{3fdc553e580f2a78}{100pt}
\inkimg{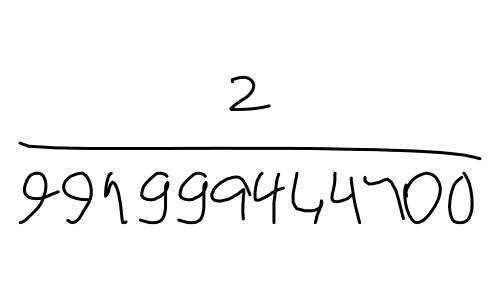}{b17c3206c2d610b9}{100pt}
\inkimg{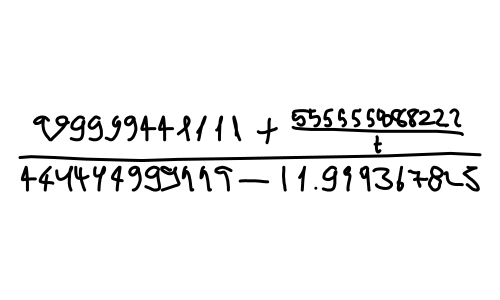}{9df33845897752ea}{100pt}
\end{center}
\section{Dataset split}\label{sec:dataset_split}
The \valid{} and \test{} splits are the result of multiple operations performed between 2016 and 2019. The first split operation, performed on the data available in 2016, was based on the contributor id: any given contributor's samples would not appear in more than one split (either \train, \valid, \test). This is common practice for handwriting recognition systems, to test how the recognizer performs on unseen handwriting styles.

Experiments then showed that a more important factor than the handwriting style was whether the \textit{label} had already been seen during training. Subsequent data collection campaigns focused on increasing label variety, and new samples were added to \valid{} and \test, this time split by label: a given normalized mathematical expression would not appear in more than one split.

\section{Label Normalization}\label{sec:appendix_normalization}

\begin{figure*}
\begin{center}

\begin{tabular}{ll}
\hline\\[-2pt]
\multicolumn{2}{l}{Sub/superscript are put in braces, \texttt{\textbackslash over} is replaced by \texttt{\textbackslash frac}} \\[4pt]
Raw: & \verb|\overline{hu^2}+{1 \over 2}{k_{ap}g_zh^2}| \\
Normalized: & \verb|\overline{hu^{2}}+\frac{1}{2}k_{ap}g_{z}h^{2}| 
\\[6pt] \hline\\[-2pt]

\multicolumn{2}{l}{Subscripts are put before superscripts, extra space is dropped} \\[4pt]
Raw: & \verb|\int^a_{-a}f(x) dx=0| \\
Normalized: & \verb|\int_{-a}^{a}f(x)dx=0| 
\\[6pt] \hline\\[-2pt]

\multicolumn{2}{l}{Single quotes are replaced by a superscript} \\[4pt]
Raw: & \verb|f'(\overline x)| \\
Normalized: & \verb|f^{\prime}(\overline{x})| 
\\[6pt] \hline\\[-2pt]

\multicolumn{2}{l}{Text formatting commands like \texttt{\textbackslash rm} are dropped} \\[4pt]
Raw: & \verb|~A_{0}=\frac{ND}{\sigma_{\rm as}+\sigma_{\rm es}}~| \\
Normalized: & \verb|A_{0}=\frac{ND}{\sigma_{as}+\sigma_{es}}| 
\\[6pt] \hline\\[-2pt]

\multicolumn{2}{l}{\parbox{14cm}{Matrix environments with delimiters like \texttt{bmatrix} are replaced by \texttt{matrix} surrounded by delimiters\\ Commands like \texttt{\textbackslash cos} are replaced by the series of letters}} \\[11pt]
Raw: & \verb|\begin{bmatrix} -\sin t \\ \cos t \end{bmatrix}| \\
Normalized: & \verb|[\begin{matrix}-sint\\ cost\end{matrix}]| 
\\[6pt] \hline\\[-2pt]

\multicolumn{2}{l}{Delimiter size modifiers like \texttt{\textbackslash big} are dropped} \\[4pt]
Raw: & \verb|\big(\tfrac{a}{N}\big)| \\
Normalized: & \verb|(\frac{a}{N})|
\\[6pt] \hline\\[-2pt]

\end{tabular}
\end{center}
\caption{Examples of expression normalization. See Section \ref{sec:normalization} for details.}
\label{fig:normalized-labels}
\end{figure*}

\subsection{Syntactic Variations}

There are several ways to change a \LaTeX{} string without changing the rendered output significantly. The normalization we implemented does the following:

\begin{itemize}
    \item all unnecessary space is dropped
    \item all command arguments are consistently put in curly braces
    \item superscripts and subscripts are put in curly braces and their order is normalized. e.g. \verb|a^2_1| becomes \verb|a_{1}^{2}|.
    \item redundant braces are dropped
    \item infix commands are replaced by their prefix versions. e.g. \verb"\over" is replaced by \verb"\frac"
    \item a lot of synonyms are collapsed. e.g. \verb"\le" and \verb"\leq", 
    \verb"\longrightarrow" and \verb"\rightarrow", etc. Some of the synonyms are only synonyms in handwriting. For example \verb"\star" ($\star$) and $*$ are different in print (5-prong and 6-prong stars), but the difference was not expressed in handwriting by our contributors.
    \item functions commands like \verb"\sin" are replaced by the sequence of letters of the function name (e.g. \verb"\sin" is replaced by \verb"sin"). This reduces the output vocabulary, and eliminates a source of confusion because we found that \LaTeX{} expressions from Wikipedia come with a mix of function commands and sequences of letters.
    \item expansion of abbreviations. e.g. \verb"\cdots", \verb"\ldots", etc. have been replaced by the corresponding sequence of characters.
    \item matrix environments are normalized to use only the 'matrix' environment surrounded by the proper delimiters like brackets or parentheses.
    \item \verb|\binom| is turned into a 2-element column matrix. Expressions from Wikipedia did not use those consistently, so we made the choice to normalize \verb|\binom| away.
\end{itemize}

\subsection{Differences Between Print And Handwriting}

The following characteristics can not be represented in handwriting and have been normalized away:

\begin{itemize}
\item color
\item accurate spacing: e.g. \verb|~|, \verb|\quad|.
\item font style and size: e.g. \verb|\mathrm|, \verb|\mathit|, \verb|\mathbf|, \verb|\scriptstyle|.
\end{itemize}

There are others that can be represented in handwriting, but that are not consistent enough in MathWriting to be preserved:

\begin{itemize}
\item font families: Fraktur, Calligraphic. In practice, only Blackboard (\verb"\mathbb") has been written consistently enough by contributors that we were able to keep it: \verb"\mathcal" and \verb"\mathfrak" are dropped.
\item some variations like \verb"\rightarrow" $\rightarrow$ and   \verb"\longrightarrow" $\longrightarrow$.
\item some character variations. e.g. \verb|\varrho|, \verb|\varepsilon|
\item size modifiers like \verb"\left", \verb"\right", \verb"\big". Similarly, variable-width diacritics like \verb"\widehat".
\end{itemize}

\section{Tokenization Code}\label{sec:tokenization}
Python code used in this work to tokenize \LaTeX{} mathematical expressions.

\begin{verbatim}
%\begin{verbbox}
import re

_COMMAND_RE = re.compile(
  r'\\(mathbb{[a-zA-Z]}|begin{[a-z]+}|end{[a-z]+}|operatorname\*|[a-zA-Z]+|.)')

def tokenize_expression(s: str) -> list[str]:
  tokens = []
  while s:
    if s[0] == '\\':
      tokens.append(_COMMAND_RE.match(s).group(0))
    else:
      tokens.append(s[0])

    s = s[len(tokens[-1]):]

  return tokens
\end{verbatim}

\end{document}